\pdfoutput=1

\documentclass[11pt]{article}

\usepackage[final]{acl}

\usepackage{times}
\usepackage{latexsym}

\usepackage[T1]{fontenc}

\usepackage[utf8]{inputenc}

\usepackage{microtype}

\usepackage{inconsolata}

\usepackage{graphicx}

\usepackage{amsmath}
\usepackage{booktabs}
\usepackage{multirow}
\usepackage{subcaption}

\title{Hopping Too Late: Exploring the Limitations of \\ Large Language Models on Multi-Hop Queries}

\author{
\textbf{Eden Biran\textsuperscript{1}}~~
\textbf{Daniela Gottesman\textsuperscript{1}}~~
\textbf{Sohee Yang\textsuperscript{2}}~~
\textbf{Mor Geva\textsuperscript{1}}~~
\textbf{Amir Globerson\textsuperscript{1,3}}
\vspace{5pt}\\
\textsuperscript{1}Tel Aviv University~~~~
\textsuperscript{2}UCL~~~~
\textsuperscript{3}Google Research
\vspace{5pt}\\
\small{\texttt{edenbiran@mail.tau.ac.il}}
}

\begin{document}
\maketitle

\begin{abstract}

Large language models (LLMs) can solve complex multi-step problems, but little is known about how these computations are implemented internally. Motivated by this, we study how LLMs answer multi-hop queries such as \emph{``The spouse of the performer of Imagine is''}.
These queries require two information extraction steps: a latent one for resolving the first hop (\emph{``the performer of Imagine''}) into the bridge entity (John Lennon), and another for resolving the second hop (\emph{``the spouse of John Lennon''}) into the target entity (Yoko Ono).
Understanding how the latent step is computed internally is key to understanding the overall computation. By carefully analyzing the internal computations of transformer-based LLMs, we discover that the bridge entity is resolved in the early layers of the model. Then, \emph{only} after this resolution, the two-hop query is solved in the later layers. 
Because the second hop commences in later layers, there could be cases where these layers no longer encode the necessary knowledge for correctly predicting the answer.
Motivated by this, we propose a novel ``back-patching'' analysis method whereby a hidden representation from a later layer is patched back to an earlier layer. We find that in up to 66\% of previously incorrect cases there exists a back-patch that results in the correct generation of the answer, showing that the later layers indeed sometimes lack the needed functionality. 
Overall, our methods and findings open further opportunities for understanding and improving latent reasoning in transformer-based LLMs.

\end{abstract}
\section{Introduction}

\begin{figure}[t]
\setlength{\belowcaptionskip}{-10pt}
\centering 
\includegraphics[scale=0.47]{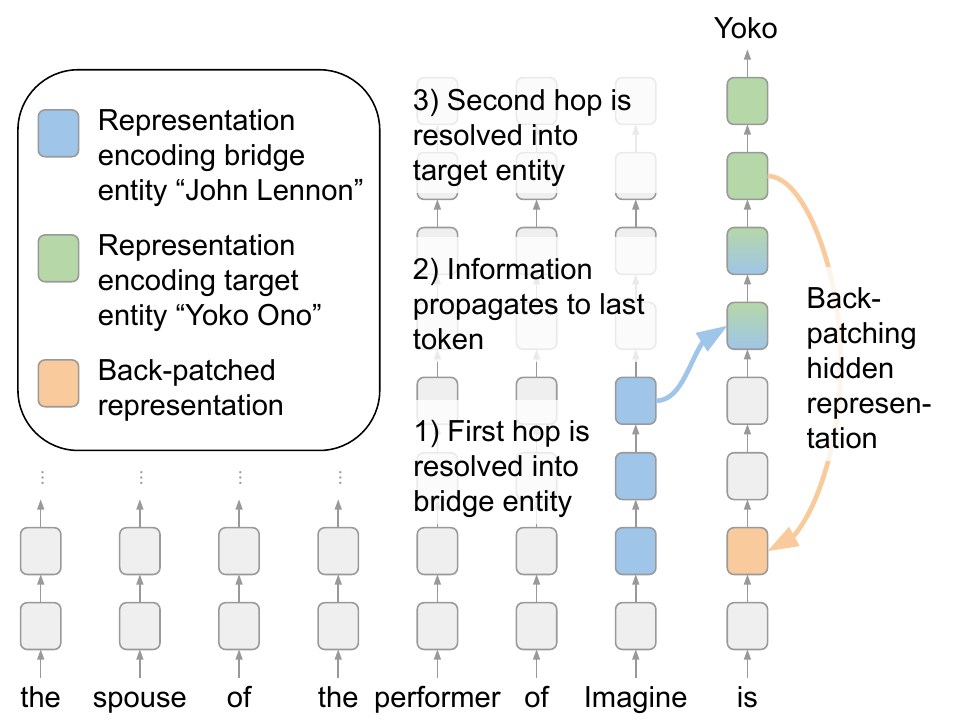}
\caption{An illustration of our findings: we observe evidence of latent reasoning in two-hop queries where 1) During the early layers the first hop is resolved and the source entity Imagine now encodes the bridge entity John Lennon 2) During the middle layers critical information propagates to the last position 3) During the later layers the second hop is resolved and the last token now encodes the target entity Yoko Ono. We additionally illustrate back-patching: patching a hidden representation from a later layer back into an earlier layer in order to fix cases where the pathway fails.}    
\label{fig:intro}
\end{figure}

Despite groundbreaking performance in a multitude of tasks, large language models (LLMs) still struggle with complex knowledge queries \citep{press-etal-2023-measuring, NEURIPS2023_deb3c281}. For example, LLMs often err when asked to complete multi-hop queries such as \emph{``The spouse of the performer of Imagine is''}. Answering such queries requires composition and reasoning skills, which have been the focus of many recent works \cite{hou-etal-2023-towards, brinkmann-etal-2024-mechanistic, yang-etal-2024-large-language-models, li-etal-2024-understanding, wang2024grokked, petty-etal-2024-impact}. Besides the general interest in compositional skills, the investigation of latent multi-hop abilities of LLMs would also have significant implications for areas such as generalization \citep{pmlr-v80-lake18a, onoe-etal-2023-lms} and model editing \citep{zhong-etal-2023-mquake, cohen-etal-2024-evaluating}.

To strengthen the latent reasoning ability of LLMs, it is essential to first understand the internal mechanism of how LLMs successfully complete two-hop queries using latent multi-hop reasoning. While there have been several works that investigate such a mechanism, a precise latent reasoning pathway has only been found in relatively small language models trained on synthetic datasets \citep{brinkmann-etal-2024-mechanistic, li-etal-2024-understanding, wang2024grokked} and has not been thoroughly investigated in large pretrained models. 

Throughout this work we consider and compare two complementing settings, one where the model correctly answers the two-hop query and the other when it does not. To this end, in \S\ref{sec:experimental_setup}, we start by creating and publishing a dataset containing 82,020 two-hop queries based on data from Wikidata \citep{vrandevcic2014wikidata}.

Our approach to analyzing multi-hop reasoning begins with the hypothesis that it involves activating the same ``knowledge-extraction module'' twice: once for the first hop (\emph{``the performer of Imagine is John Lennon''}) and once for the second (\emph{``the spouse of John Lennon is Yoko Ono''}). The key question we ask in this work is where are these two procedures implemented in the LLM. A natural way to answer this question is to seek the first point in the network where the outputs of these two steps appear. The typical approach to this would employ vocabulary projections \citep{nostalgebraist2020, geva-etal-2021-transformer, sakarvadia-etal-2023-memory, yang-etal-2024-large-language-models, li-etal-2024-understanding}. However, we instead opt for a more recent method named Patchscopes \citep{ghandeharioun2024patchscopes} which has clear advantages over the commonly used vocabulary projection. 

Using this approach, in \S\ref{sec:first_hop_resolved} we find that the outcome of the first hop is clearly evident in the hidden representation of the end-token of the first hop. We further find that resolving the first hop happens during the early layers, as depicted in Figure \ref{fig:intro} (stage 1). Given the localization of the first hop, in \S\ref{sec:second_hop_resolved} we then ask where the second hop is resolved. We find that it typically appears in the last token of the prompt \emph{only} in layers after the resolution of the first hop, as further illustrated in Figure \ref{fig:intro} (stage 3).

These observations suggest a mechanism where the first hop is resolved with the answer, that then propagates to the last token to resolve the second hop (as depicted by stage 2 in Figure \ref{fig:intro}). We explore this propagation in \S\ref{sec:info_propagates}. Moreover, they suggest why this process may fail --- If the two resolutions are done by different groups of layers, it must be that these two groups are both able to perform knowledge extraction, be it via their attention or MLP sublayers.
However, since transformers have a limited amount of layers, and their functionality is different \citep{NEURIPS2022_6f1d43d5, geva-etal-2023-dissecting}, it is quite likely that the layer at which the second hop is performed will not have the desired functionality. To verify this hypothesis, in \S\ref{sec:backpatching}, we propose an analysis method named \textit{back-patching}. The crux of the method involves taking a hidden representation from a later layer, patching it back into the same position of the same prompt in an earlier layer, and then continuing the forward pass. In a way, this allows the model more layers to finish the computation without the need for additional parameters or training \citep{petty-etal-2024-impact}. Unlike previous methods \citep{sakarvadia-etal-2023-memory, li-etal-2024-understanding}, back-patching makes no assumption on the structure of the prompt and does not require any knowledge of the participating entities. Testing back-patching on queries initially predicted incorrectly, we find that up to 66\% of incorrect cases have the potential to be correctly predicted using back-patching when choosing the optimal source and target layers. 

Overall, our contributions can be summarized as follows:
\begin{itemize}
  \item We provide a novel dataset of two-hop queries, which can serve to systematically study this important setting.
  \item We employ Patchscopes to inspect entities encoded in hidden representations while processing two-hop queries.
  \item We identify a sequential latent reasoning pathway in LLMs, where the first hop query is initially resolved into the bridge entity which is then used to answer the second hop.
  \item We propose an analysis method named back-patching, that verifies our results and could additionally be used to improve the performance of LLMs on multi-hop question answering.
\end{itemize}

We release our code and dataset at \url{https://github.com/edenbiran/HoppingTooLate}.

\section{Experimental Setup} \label{sec:experimental_setup}

\subsection{Two-Hop Queries}
We consider facts denoted by a triplet $(e, r, e')$ where $e$ is a source entity (e.g., John Lennon), $r$ is a relation (e.g., Spouse) and $e'$ is a target entity (e.g., Yoko Ono). These facts can then be converted to factual statements (e.g., \emph{``The spouse of John Lennon is Yoko Ono''}) and also into queries by omitting $e'$ (e.g., \emph{``The spouse of John Lennon is ''}). We then prompt models with these queries in order to test their knowledge and reasoning.

We create two-hop queries by composing two facts where one's target entity is the other's source entity. Namely, $((e_1, r_1, e_2)$ and $(e_2, r_2, e_3))$. We refer to $e_1$ as the source entity, $e_2$ as the bridge entity, and $e_3$ as the target entity. For example, the two-hop query \emph{``The spouse of the performer of Imagine is''} can be decomposed into (Imagine, Performer, John Lennon) and (John Lennon, Spouse, Yoko Ono). 

We further define two tokens of specific interest: the last token of $e_1$\footnote{This token is also the last token of the first hop.} (e.g., \emph{``The spouse of the performer of \underline{Imagine}''}) and the last token of the whole two-hop prompt (e.g., \emph{``The spouse of the performer of Imagine \underline{is}''}), which we denote as $t_1$ and $t_2$, respectively.

\subsection{Dataset}

We create a dataset containing 82,020 two-hop queries based on data from Wikidata \citep{vrandevcic2014wikidata}. First, we sample entities with a maximal amount of Wikidata statements from a set of predefined Wikidata entity types. These entities will act as $e_2$. Second, we use a predefined set of relations in order to construct the two triplets $(e_1, r_1, e_2)$ and $(e_2, r_2, e_3)$. Finally we convert both triplets into a single natural language phrase using manually crafted templates per relation. We refer to our codebase for further details including entity types and relations\footnote{All data was retrieved during April 2024.}.

We next attempt to filter out cases where no latent reasoning is performed. Given a two hop query $((e_1, r_1, e_2), (e_2, r_2, e_3))$ we test two prompts constructed to detect and filter out cases where the model performs reasoning shortcuts \citep{xu-etal-2022-model, wang-etal-2023-causal, ju-etal-2024-investigating}. The first prompt is the query $((\text{""}, r_1, e_2), (e_2, r_2, e_3))$ (i.e., the query without $e_1$), aimed at filtering out cases where the model predicts generally popular entities. The second prompt is $((e_1, \text{""}, e_2), (e_2, r_2, e_3))$ (i.e., the query without $r_1$), aimed at filtering out cases where the model predicts correctly due to a high correlation between $e_1$ and $e_3$. For example, given the two-hop query \emph{``The spouse of the performer of Imagine is''} we filter out cases where the model predicts Yoko Ono for either \emph{``The spouse of the performer is''} or \emph{``The spouse of Imagine is''}. We perform this filtering for each model using greedy decoding creating a per model subset of the dataset. Table \ref{tab:cases_counts} displays the amount of examples left after this filtering process.

\begin{table}[t]
\centering
\setlength{\belowcaptionskip}{-13pt}
\small
\begin{tabular}{lccc}
    \toprule
    \multirow{2}{*}{Model} & \multirow{2}{*}{Post Filtering} & \multicolumn{2}{c}{Cases Tested} \\ 
    & & Correct & Incorrect \\ \midrule
    LLaMA 2 7B & $70,625$ & $388$ & $351$\\ \midrule
    LLaMA 2 13B & $70,972$ & $554$ & $413$ \\ \midrule
    LLaMA 3 8B & $71,569$ & $379$ & $371$ \\ \midrule
    LLaMA 3 70B & $70,334$ & $656$ & $595$ \\ \midrule
    Pythia 6.9B & $73,058$ & $173$ & $202$ \\ \midrule
    Pythia 12B & $74,056$ & $172$ & $372$ \\
    \bottomrule
\end{tabular}
\caption{The amount of examples post shortcut filtering and the final amount used throughout all experiments.}
\label{tab:cases_counts}
\end{table}

We are specifically interested in understanding the differences between cases where the model completes the two-hop query correctly and incorrectly. Therefore we generate two dataset subsets per model for these cases accordingly. 

The first subset is made up of cases where the model correctly answers both the two-hop query and the first hop. For example, given the two-hop query \emph{``The spouse of the performer of Imagine is''}, we verify that the model correctly predicts the answer Yoko Ono when prompted with the two-hop query and additionally correctly predicts John Lennon when prompted with the first hop \emph{``The performer of Imagine is''}. We use the latter filter to make sure that the model indeed ``knows'' the answer to the first hop. 
We then randomly sample 100 examples of each bridge entity type, in order to create a more balanced subset (if less than 100 examples of a specific type exist we take them all). 

The second subset includes cases where the model correctly answers both the first and second hop in isolation, but fails to answer the full two-hop query. As an example for the filtering procedure, given the same query above, we verify that the model correctly predicts John Lennon for the first hop \emph{``The performer of Imagine is''} and Yoko Ono for the second hop \emph{``The spouse of John Lennon is''}, but fails to predict Yoko Ono for the two-hop query \emph{``The spouse of the performer of Imagine is''}. We focus on these cases, because we are interested in understanding why answering the two-hop query fails despite the model ``knowing'' the two separate facts. As we show later, this failure is likely due to the first fact being resolved too late. In a similar fashion to the first setting, we then randomly sample 50 examples of each bridge entity type (in order to keep the correct and incorrect subsets of approximately the same size). 

Table \ref{tab:cases_counts} reports the amount of examples used in experiments by model and setting.

\subsection{Models}

We analyze the following models: LLaMA 2 7B and 13B \citep{touvron2023llama}, LLaMA 3 8B and 70B \citep{llama3}, and Pythia 6.9B and 12B \citep{biderman2023pythia}. The 6.9B, 7B and 8B models have 32 layers, the 12B model has 36 layers, the 13B model has 40 layers, and the 70B model has 80 layers.

\section{Localizing First Hop Resolution} \label{sec:first_hop_resolved}

One could imagine several possible ways for a model to solve multi-hop queries. One such way would be to effectively treat the two-hop query as a single relation, enabling the extraction of the final answer without the need to recall $e_2$. 
Despite this, if the model truly performs latent reasoning, the most intuitive way to do so would start by resolving the first hop at some point within the computation. In order to localize where this happens, we use the Patchscopes analysis method. Our results show that the model indeed resolves the first hop during the early layers, at the position of $t_1$. 

\subsection{Interpreting Hidden Representations} 

We use the Patchscopes framework \citep{ghandeharioun2024patchscopes} to create a task that describes the entity encoded in a specific hidden representation. Patchscopes maps a given representation to a sentence in natural language, thus considerably extending vocabulary projections which map to a single token \citep{nostalgebraist2020}. 

The procedure is as follows. First, given a source prompt, a source token and a source layer, the source prompt is passed through the model's forward computation and the hidden representation $v$ of the source token at the source layer is recorded. This representation $v$ is what we aim to probe in search for an encoded entity. Second, we pass the same prompt used by \citet{ghandeharioun2024patchscopes}: \texttt{``Syria: Syria is a country in the Middle East, Leonardo DiCaprio: Leonardo DiCaprio is an American actor, Samsung: Samsung is a South Korean multinational corporation, x''} through the model, replacing the hidden representation of \texttt{``x''} with $v$ at a specific target layer. Finally, the forward pass is continued and text is generated.

The chosen target prompt encourages the model to generate a continuation that states the semantic content (i.e., the entity name) encoded in the source hidden representation, along with a short description of it. We find this task is better fitting than other proposed probes such as vocabulary projections \citep{nostalgebraist2020, geva-etal-2022-transformer} or training linear classifiers on hidden representations \citep{belinkov-glass-2019-analysis, belinkov-2022-probing}. This is because exhibiting the ability to extract information from a hidden representation gives reason to believe that the same information can be extracted when answering the two-hop query. In addition Patchscopes has the clear advantage of decoding a representation into a natural language description.

\subsection{Experiment} 
We run Patchscopes on $t_1$, recording the hidden representation at each source layer and then patching it into all target layers. Following each patch we sample three generations. For each source layer we check whether one of the generations is of $e_2$, and if so, we consider the representation to encode the bridge entity at the source layer.

\subsection{Results} 

\begin{table}[t]
\centering
\small
\setlength\tabcolsep{2pt}
\begin{tabular}{llcccccc}
    \toprule
    \multirow{2}{*}{Model} & \multirow{2}{*}{Subset} & \multicolumn{2}{c}{$e_2$ from $t_1$} & \multicolumn{2}{c}{$e_2$ from $t_2$} & \multicolumn{2}{c}{$e_3$ from $t_2$} \\ 
    & & Cases & Layer & Cases & Layer & Cases & Layer \\ \midrule
    LLaMA 2 & Corr. & $56\%$ & $8.9$ & $41\%$ & $15.3$ & $73\%$ & $16.2$ \\
    7B & Incorr. & $41\%$ & $9.1$ & $36\%$ & $16.4$ & $47\%$ & $17.5$ \\ \midrule
    LLaMA 2 & Corr. & $49\%$ & $7.7$ & $34\%$ & $18.2$ & $71\%$ & $16.9$ \\
    13B & Incorr. & $48\%$ & $7.4$ & $37\%$ & $17.9$ & $33\%$ & $16.9$ \\ \midrule
    LLaMA 3 & Corr. & $46\%$ & $8.9$ & $37\%$ & $14.3$ & $74\%$ & $13.5$ \\
    8B & Incorr. & $46\%$ & $11.9$ & $25\%$ & $15.3$ & $40\%$ & $17.2$ \\ \midrule
    LLaMA 3 & Corr. & $63\%$ & $27.1$ & $46\%$ & $35.2$ & $86\%$ & $30.7$ \\
    70B & Incorr. & $57\%$ & $29.2$ & $46\%$ & $34.5$ & $50\%$ & $35.9$ \\ \midrule
    Pythia & Corr. & $75\%$ & $5.4$ & $75\%$ & $11.4$ & $80\%$ & $14.1$ \\
    6.9B & Incorr. & $78\%$ & $4.9$ & $67\%$ & $9.2$ & $65\%$ & $13.6$ \\ \midrule
    Pythia & Corr. & $73\%$ & $5.0$ & $66\%$ & $12.2$ & $77\%$ & $13.5$ \\
    12B & Incorr. & $61\%$ & $6.3$ & $42\%$ & $12.9$ & $52\%$ & $17.2$ \\
    \bottomrule
\end{tabular}
\caption{The results of Patchscopes executions. The table reports the percentage of cases where a target entity was successfully decoded from a source position. The table additionally reports the mean of the layer where the target entity was first successfully decoded.}
\label{tab:entity_description}
\end{table}

\begin{figure}[t]
\centering 
\includegraphics[scale=0.5]{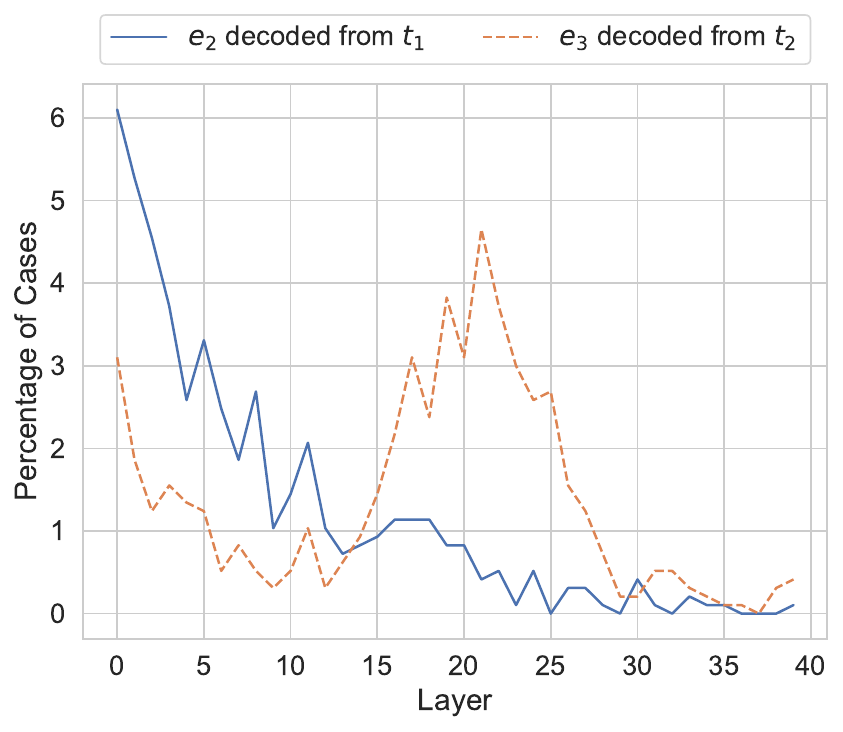}
\caption{Percentage of cases per layer where target entities were first successfully decoded using Patchscopes. The percentages are out of all correctly answered cases for LLaMA 2 13B.}    
\label{fig:min_entity_description_layers_llama2-13b}
\end{figure}

\begin{figure}[t]
\centering 
\includegraphics[scale=0.12]{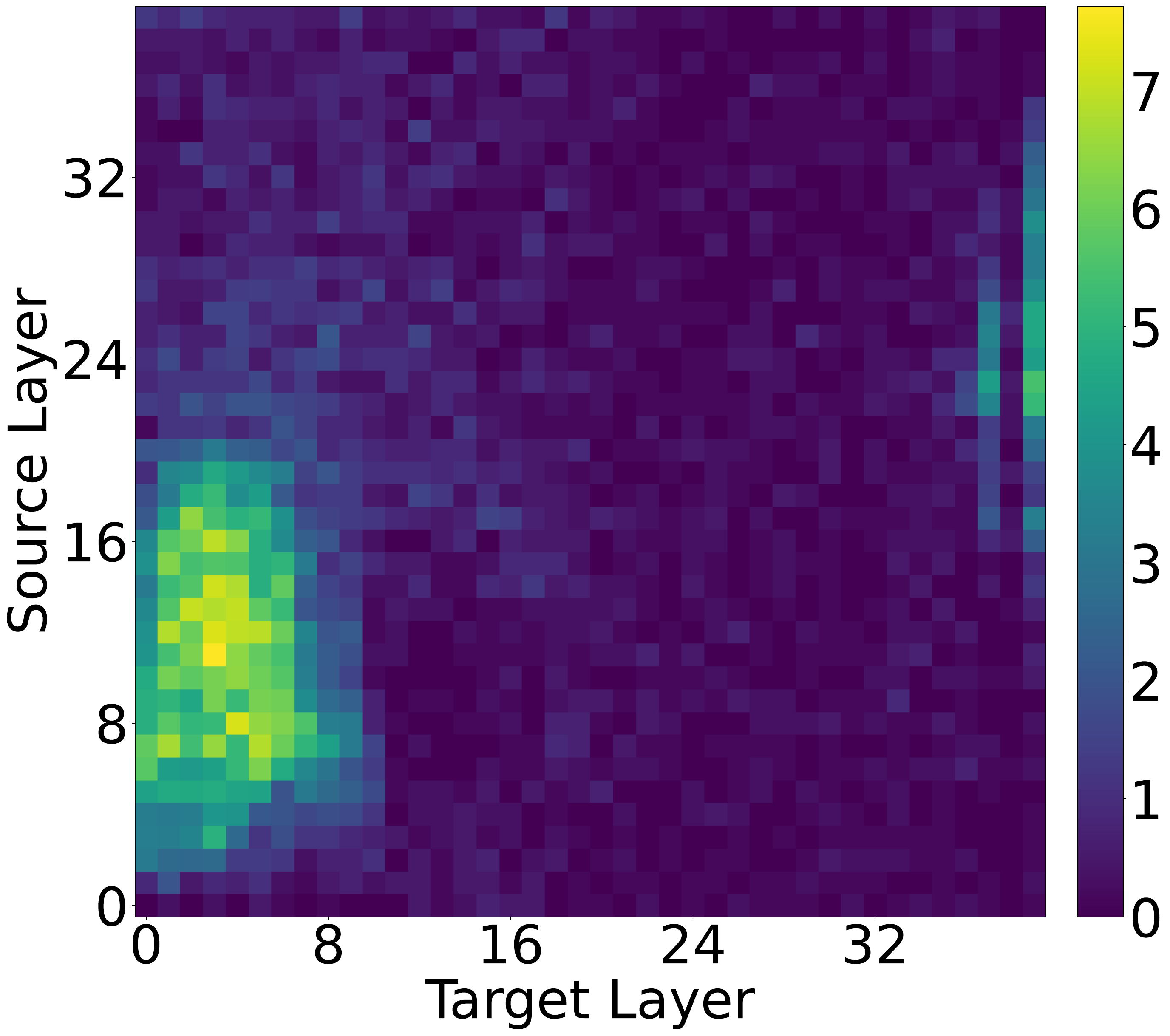}
\caption{Heat-map of the layers where Patchscopes successfully decodes $e_2$ from the position of $t_1$. The percentages are out of all successful decodings for LLaMA 2 13B run on correctly answered cases.}
\label{fig:entity_description_e1_e2_matrix_llama2-13b}
\end{figure}

Table \ref{tab:entity_description} presents the percentage of cases where $e_2$ was observed in the hidden representation of $t_1$. We refer to these as cases where $e_2$ was successfully resolved. We note the non-trivial percentages of 41\%-78\% across all models and settings. Additionally, we note the general drop in resolved cases when the model fails to correctly complete the two-hop query, but that this setting still displays a surprisingly high success rate. Next, we examine the mean layer at which $e_2$ first appears in both correct and incorrect cases. We find that in the incorrect cases the resolving generally occurs at higher layers than those of the correct cases. This difference alludes to how early the resolving happens potentially playing a part in successfully answering two-hop queries. We further verify this using the back-patching analysis method in \S\ref{sec:backpatching}.

Figure \ref{fig:min_entity_description_layers_llama2-13b} displays the proportion of cases per layer where $e_2$ is first successfully decoded from $t_1$ using Patchscopes. We observe that the resolving occurs almost exclusively during the early layers. Interestingly, this implies that the knowledge required for this resolution must reside in these earlier layers. Similar figures for additional models are provided in Appendix \ref{app:ent_desc_decoded_layers}. The plot additionally shows the curve regarding the resolving of $e_3$ from $t_2$, which is discussed in \S\ref{sec:second_hop_resolved}.

This finding can further be seen in Figure \ref{fig:entity_description_e1_e2_matrix_llama2-13b}, which displays a heat-map of the layers where Patchscopes successfully decodes $e_2$ from $t_1$. Two hot-spots are apparent. The first corresponds to success cases with an early source layer, showing that $e_2$ generally appears at this position during the early layers. The second hot-spot corresponds to later source layers that display success almost exclusively when patched after the final layer, which is roughly equivalent to vocabulary projection \citep{nostalgebraist2020}. This is a clear advantage of Patchscopes, as despite the entity not being among the top tokens in the projection during early layers, it is in fact encoded in the hidden representation. See Appendix \ref{app:ent_desc_heatmaps} for heat-maps of all models.
\section{Second Hop is Resolved at Last Position} \label{sec:second_hop_resolved}

In \S\ref{sec:first_hop_resolved}, we observe that the first hop is often resolved in position $t_1$. Intuitively, one would expect the computation to use this result towards resolving the second hop (i.e., the final answer). In this section, we study this proposition by finding where $e_3$ first emerges and how it relates to the first emergence of $e_2$. We find that $e_3$ indeed emerges predominantly after $e_2$ at the position of $t_2$. Furthermore, we search for the final part of the pathway, the predicted token. By projecting the residual updates made by the attention and MLP sublayers on the hidden representation of $t_2$, we find that the MLP sublayers play the larger part in conjuring the predicted token.

\subsection{Method}

Previous works \citep{geva-etal-2022-transformer, geva-etal-2023-dissecting} have shown that the residual updates made by the MLP and attention sublayers can be interpreted by projecting them to the vocabulary using the layer norm and output embedding matrix. We use this method in order to locate specific entities in the flow of information by observing the rank of the entity in the vocabulary projection. We additionally use Patchscopes as presented in \S\ref{sec:first_hop_resolved}.

\subsection{Experiment}

We first employ Patchscopes on $t_2$, this time checking if the target entity $e_3$ is encoded at this position. Additionally, we project the sublayer updates performed on $t_2$, and check if the token with the highest probability in the projection is the first token of the generated text. If so, we conclude that the update of a particular sublayer promotes the final prediction.

\subsection{Results}

\begin{table}[t]
\setlength\tabcolsep{3pt}
\centering
\small
\begin{tabular}{llcccc}
    \toprule
    \multirow{2}{*}{Model} & \multirow{2}{*}{Subset} & \multicolumn{2}{c}{Attention} & \multicolumn{2}{c}{MLP} \\ 
    & & Cases & Layer & Cases & Layer \\ \midrule
    \multirow{2}{*}{LLaMA 2 7B} 
    & Correct & $25.5\%$ & $24.7$ & $33.2\%$ & $25.2$ \\
    & Incorrect & $14.8\%$ & $23.4$ & $28.4\%$ & $26.0$\\ \midrule
    \multirow{2}{*}{LLaMA 2 13B} 
    & Correct & $28.8\%$ & $32.0$ & $68.5\%$ & $33.1$ \\
    & Incorrect & $14.2\%$ & $27.0$ & $51.5\%$ & $31.7$ \\ \midrule
    \multirow{2}{*}{LLaMA 3 8B} 
    & Correct & $9.7\%$ & $26.7$ & $17.6\%$ & $27.8$ \\
    & Incorrect & $25.3\%$ & $27.4$ & $11.8\%$ & $24.6$ \\ \midrule
    \multirow{2}{*}{LLaMA 3 70B} 
    & Correct & $21.9\%$ & $56.3$ & $48.1\%$ & $68.0$ \\
    & Incorrect & $21.8\%$ & $65.7$ & $36.1\%$ & $67.0$ \\ \midrule
    \multirow{2}{*}{Pythia 6.9B} 
    & Correct & $11.5\%$ & $21.8$ & $36.4\%$ & $20.5$ \\
    & Incorrect & $28.2\%$ & $23.4$ & $33.6\%$ & $21.2$ \\ \midrule
    \multirow{2}{*}{Pythia 12B} 
    & Correct & $27.9\%$ & $22.2$ & $38.9\%$ & $22.4$ \\
    & Incorrect & $43.2\%$ & $23.6$ & $47.0\%$ & $23.3$ \\
    \bottomrule
\end{tabular}
\caption{Percentages and mean layers where for the first time the predicted token is the most probable token in the vocabulary projection of the sublayer update.}
\label{tab:prediction_promoted}
\end{table}

Table \ref{tab:entity_description} contains the percentage of cases $e_3$ was decoded from $t_2$, effectively resolving the second hop. As expected, these percentages are high for cases in which the model predicted correctly (71\%-86\%)\footnote{We attribute the gap from 100\% to the approximations involved in decoding a hidden representation.} and low for cases the model predicted incorrectly (33\%-65\%). Additionally, it is apparent from both Table \ref{tab:entity_description} and Figure \ref{fig:min_entity_description_layers_llama2-13b} that the target entity is resolved by the mid-upper layers at the position of $t_2$. This suggests that the information required to perform the second hop exists in the parameters of these layers.

Inspecting the sublayer projections in Table \ref{tab:prediction_promoted}, we find that the MLP sublayers play a larger role in promoting the predicted token compared to that of the attention sublayers. Despite this, the non negligible percentages achieved by the attention sublayers could suggest multiple pathways, in agreement with previous work \citep{geva-etal-2023-dissecting, merullo-etal-2024-language} that considered both possibilities. Looking at the mean layers at which the prediction is first promoted, one can see that these promotions occur in the upper layers, after the last token has been resolved to $e_3$.
\section{Information Propagates to Last Token} \label{sec:info_propagates}

As the model's prediction is emitted from the last position of the prompt, critical information regarding the bridge entity must propagate to this token. By using attention knockout, projection, and Patchscopes, we verify this claim. We find that in a large majority of cases relevant information can indeed be detected passing from $t_1$ to $t_2$. 

\subsection{Method} 

In similar fashion to previous work \citep{wang2023interpretability, geva-etal-2023-dissecting}, we block hidden representations from attending to other representations at specific layers and test if the model's final generation changes. We perform the knockout by setting the mask of the attention from a source to a target position at a specific layer to $-\infty$, causing this attention edge to contribute nothing to the residual update performed on the source representation. We experiment with blocking a window of layers as previous work \citep{geva-etal-2023-dissecting} shows that information propagation is spread among multiple layers.

We additionally employ Patchscopes and sublayer vocabulary projections, as described in \S\ref{sec:first_hop_resolved} and \S\ref{sec:second_hop_resolved} respectively.

\subsection{Experiment} 

We run attention knockout blocking information originating at $t_1$ from propagating to $t_2$. We block a window of 7 layers in order to capture information flowing through multiple layers. If blocking a set of layers causes the model to no longer correctly predict the answer (in the incorrect setting we check for a change in the generated text), we consider these layers to contain information critical to the prediction. 

We additionally project the updates made by the attention and MLP sublayers to the last token, as described in \S\ref{sec:second_hop_resolved}. In the vocabulary projection we check if $e_2$ is the token with the highest probability, and if so consider the update to ``contain'' the bridge entity. 

Finally, we run Patchscopes on the last token of the prompt, as done in \S\ref{sec:first_hop_resolved}, checking if $e_2$ is encoded in $t_2$. Each one of these experiments gives a unique signal that information must have propagated from $t_1$ to $t_2$.   

\subsection{Results} 

\begin{table}[t]
\centering
\small
\begin{tabular}{llcc}
    \toprule
    Model & Subset & Detected & Mean Layer \\ \midrule
    \multirow{2}{*}{LLaMA 2 7B} 
    & Correct & $85.82\%$ & $12.83$ \\
    & Incorrect & $95.44\%$ & $8.10$ \\ \midrule
    \multirow{2}{*}{LLaMA 2 13B} 
    & Correct & $81.94\%$ & $15.45$ \\
    & Incorrect & $71.11\%$ & $16.99$ \\ \midrule
    \multirow{2}{*}{LLaMA 3 8B} 
    & Correct & $78.36\%$ & $10.15$  \\
    & Incorrect & $82.74\%$ & $7.45$ \\ \midrule
    \multirow{2}{*}{LLaMA 3 70B} 
    & Correct & $90.39\%$ & $29.14$ \\
    & Incorrect & $94.28\%$ & $20.06$ \\ \midrule
    \multirow{2}{*}{Pythia 6.9B} 
    & Correct & $96.53\%$ & $11.94$  \\
    & Incorrect & $95.04\%$ & $8.06$ \\ \midrule
    \multirow{2}{*}{Pythia 12B} 
    & Correct & $94.18\%$ & $11.23$ \\
    & Incorrect & $93.81\%$ & $9.28$ \\
    \bottomrule
\end{tabular}
\caption{Percentages and mean layers of cases where critical information was first detected propagating from $t_1$ to $t_2$ using attention knockout, projection, or Patchscopes.}
\label{tab:bridge_entity_propagates}
\end{table}

Table \ref{tab:bridge_entity_propagates} shows the percentage of cases where we detect critical information propagating from $t_1$ to $t_2$ using at least one of the methods stated above. We note the relatively high percentages, indicating the information indeed flows as expected. Additionally, the table shows the that the layers relevant to the propagation are mostly the middle layers.

Table \ref{tab:entity_description} contains the percentage of cases $e_2$ was decoded from $t_2$, showing that this position indeed represents the bridge entity in 25\%-75\% of all cases. Comparing between the results of Patchscopes on $t_2$ to those on $t_1$, we observe a stark difference in the successful source layers. The results show that the $t_2$ first encodes $e_2$ during the mid-upper layers, while $t_1$ does so in the lower layers. This can further be seen in the Patchscopes heat-maps in Appendix \ref{app:ent_desc_heatmaps}. This gives additional reason to believe that a sequential computation is taking place. 

\begin{figure}[t]
\centering 
\includegraphics[scale=0.52]{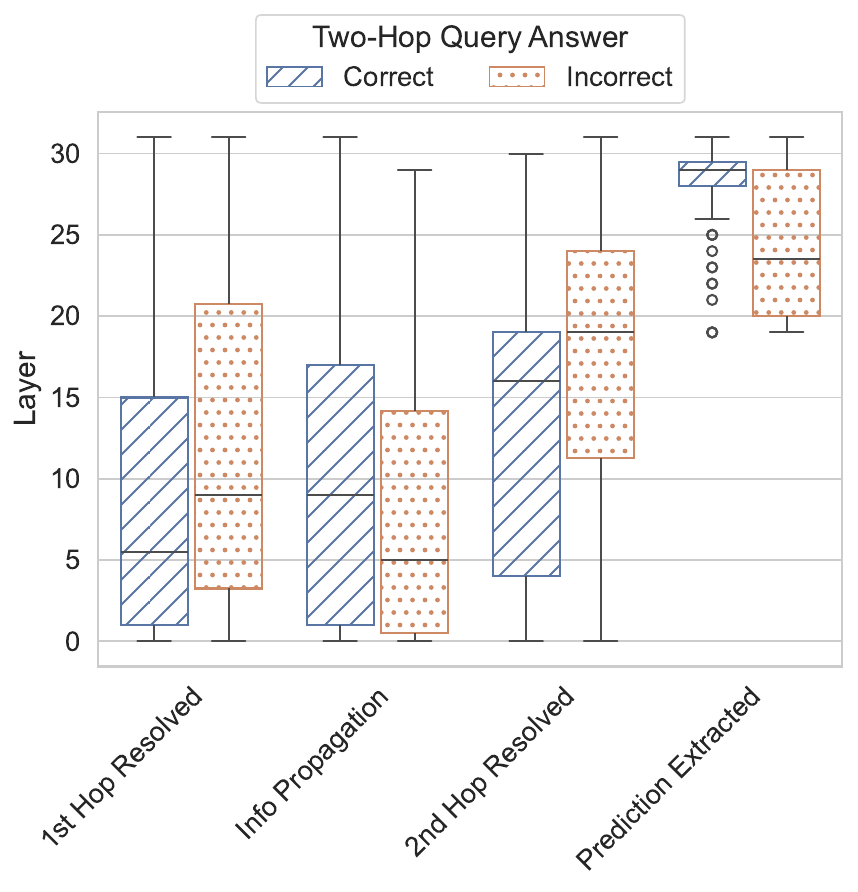}
\caption{A comparison of the first layers of each stage in the pathway between correct and incorrect cases for LLaMA 3 8B.}    
\label{fig:stages_boxplot_llama3-8b}
\end{figure}

After describing all major pathway stages, we turn to a comparison between the correct and incorrect cases and observe the following pattern. Generally, in incorrect cases, the entities seem to be resolved later while the information seems to propagate and be extracted earlier. Figure \ref{fig:stages_boxplot_llama3-8b} displays this pattern in LLaMA 3 8B, plotting the first observed layers of each stage (we refer to Appendix \ref{app:layers_of_stages} for plots of additional models). This gives evidence of the importance of how early the first hop is resolved.
\section{Back-patching Improves Two-Hop Performance} \label{sec:backpatching}

\begin{figure*}[t]
\centering
\setlength{\belowcaptionskip}{-5pt}
\begin{subfigure}{.5\textwidth}
  \centering
  \includegraphics[scale=0.12]{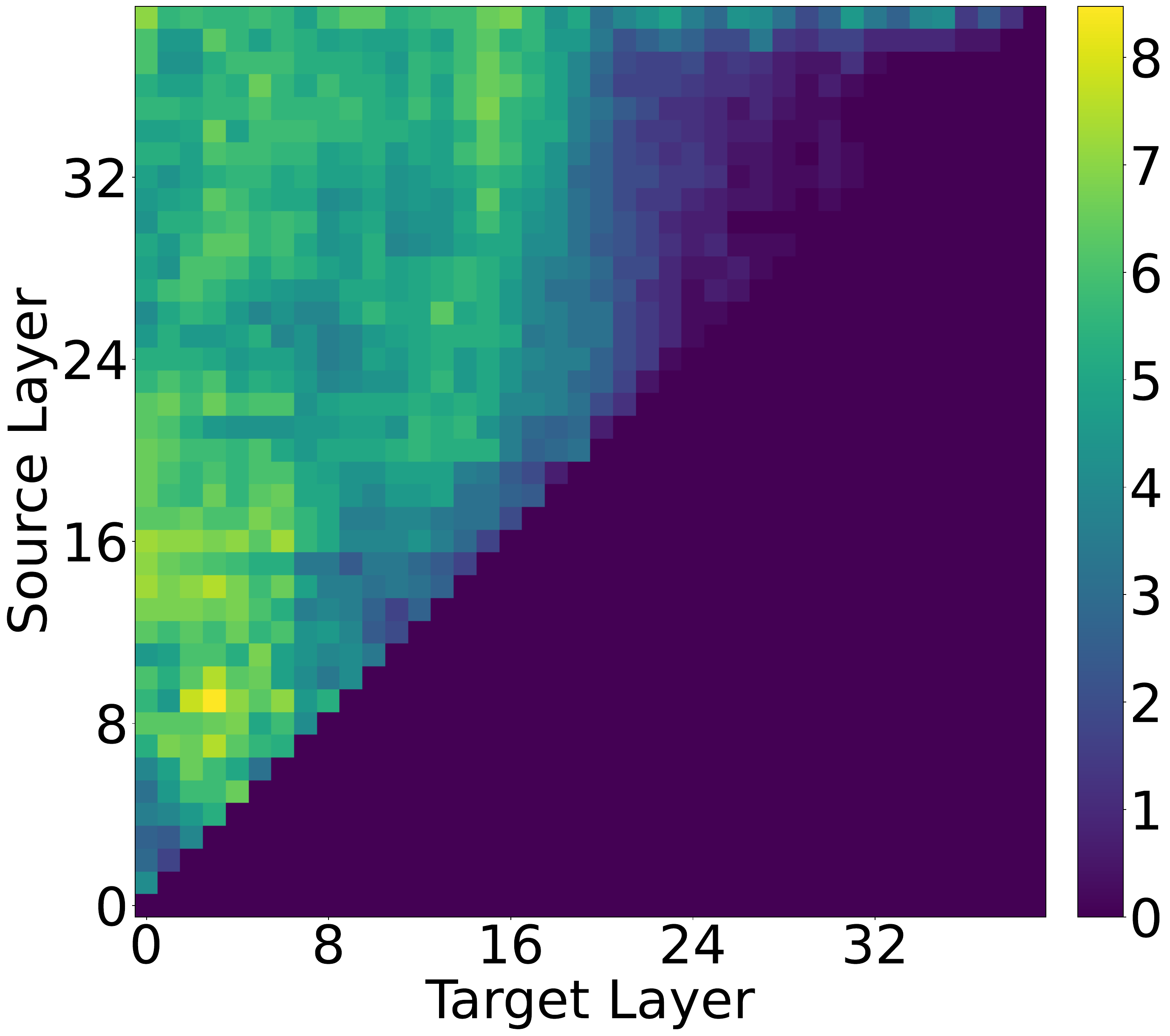}
  \caption{Back-patching $t_1$}    
  \label{fig:backpatching_e1_llama2-13b}
\end{subfigure}%
\begin{subfigure}{.5\textwidth}
  \centering
  \includegraphics[scale=0.12]{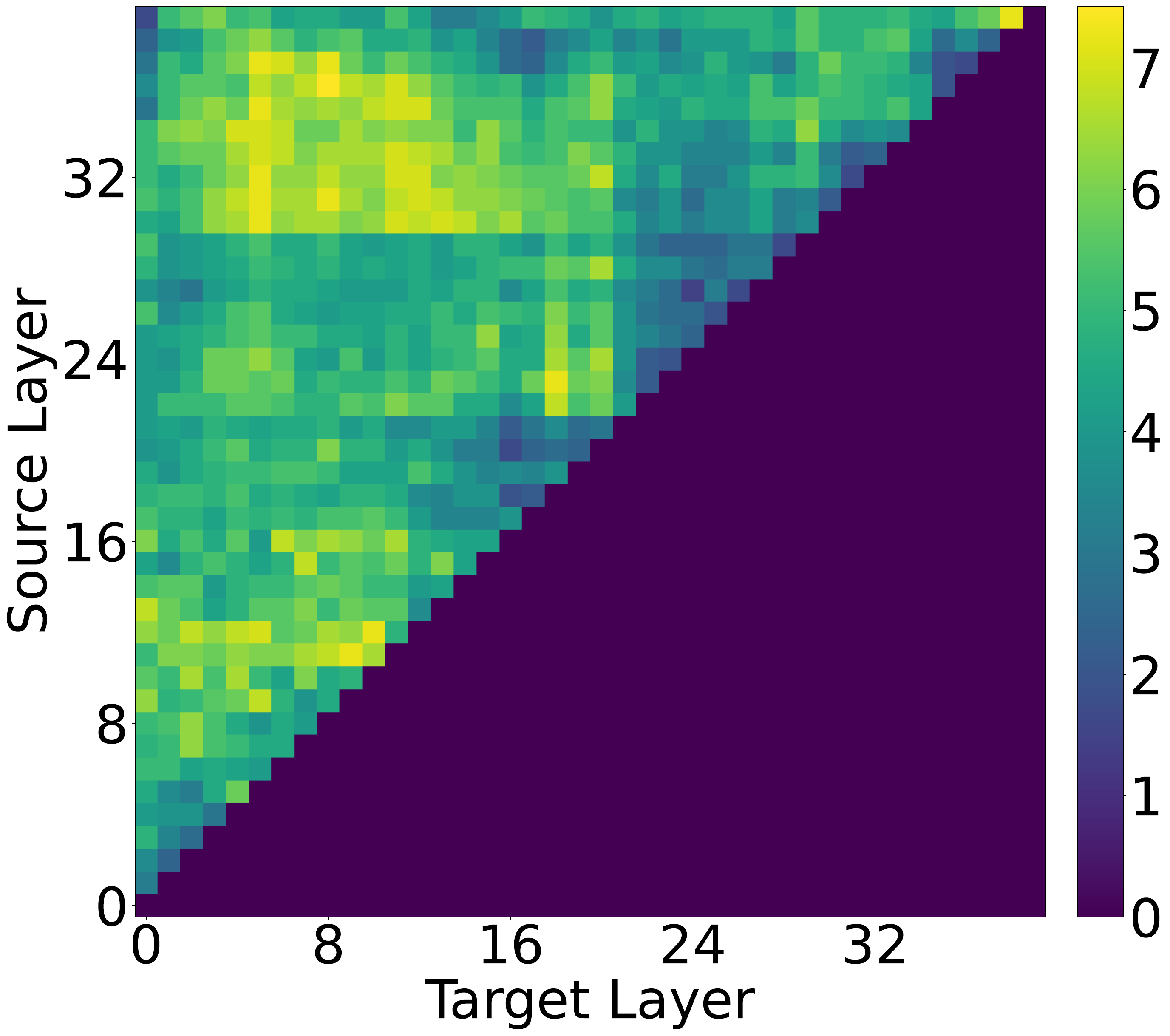}
  \caption{Back-patching $t_2$}    
  \label{fig:backpatching_last_llama2-13b}
\end{subfigure}%
\caption{Heat-map of the layers where back-patching succeeds. The percentages are out of all successful back-patching instances for LLaMA 2 13B.}    
\label{fig:backpatching_llama2-13b}
\end{figure*}

Our results consistently display a sequential nature of computation, where the first hop is resolved into the bridge entity by the lower layers, which is then used to answer the second hop in the upper layers. These results suggest that when answering two-hop queries fail, it could be because the first-hop is resolved at layers that no longer contain the information needed to resolve the second hop. In order to verify this hypothesis, we test the following intriguing conjecture. We know that the earlier layers certainly contain knowledge, since they are regularly used to resolve the first hop. Then, we ask what will happen if we take a hidden representation from a later layer and ``feed'' it into these earlier layers. Will the correct information now be extracted? As our results next show, this is often the case. This ``back-patching'' analysis thus provides strong support that the success in two hop queries require the later layers to have the same functionality as earlier ones.

Overall, this could point to an inherent limitation of the transformer architecture, where the model ``runs out'' of layers, propagates and extracts information before it is fully resolved, or is forced to use only the upper layers for knowledge extraction.

\subsection{Method}

Following our intuition above, we perform the following back-patching experiment. Given a prompt, we first record a hidden representation $v$ from a source layer $\ell_s$. Then, while rerunning the same prompt, we patch $v$ into the same position at a target layer $\ell_t$ such that $\ell_t < \ell_s$. The forward pass then continues and text is generated using greedy decoding. This effectively feeds $\ell_t$ with information encoded in $v$ that originated in $\ell_s$.

\subsection{Experiment} 

We perform the back-patching experiment twice, once on the token $t_1$ and once on the token $t_2$. In order to quantify the potential of back-patching, we check whether one of the source-target pairs resulted in the generation of the correct answer. We then perform an analysis of the successful source and target layers.

\subsection{Results} 

\begin{table}[t]
\centering
\small
\begin{tabular}{llcc}
    \toprule
    Model & Subset & $t_1$ & $t_2$ \\ \midrule
    \multirow{2}{*}{LLaMA 2 7B} 
    & Correct & $100\%$ & $100\%$ \\
    & Incorrect & $41.02\%$ & $42.45\%$ \\ \midrule
    \multirow{2}{*}{LLaMA 2 13B} 
    & Correct & $100\%$ & $100\%$ \\
    & Incorrect & $32.44\%$ & $36.07\%$ \\ \midrule
    \multirow{2}{*}{LLaMA 3 8B} 
    & Correct & $100\%$ & $100\%$ \\
    & Incorrect & $38.81\%$ & $47.16\%$ \\ \midrule
    \multirow{2}{*}{LLaMA 3 70B} 
    & Correct & $100\%$ & $100\%$ \\
    & Incorrect & $57.31\%$ & $57.81\%$ \\ \midrule
    \multirow{2}{*}{Pythia 6.9B} 
    & Correct & $100\%$ & $100\%$ \\
    & Incorrect & $66.33\%$ & $56.43\%$ \\ \midrule
    \multirow{2}{*}{Pythia 12B} 
    & Correct & $100\%$ & $100\%$ \\
    & Incorrect & $63.17\%$ & $61.82\%$ \\
    \bottomrule
\end{tabular}
\caption{Back-patching success rates by setting and back-patched token.}
\label{tab:backpatching}
\end{table}

Table \ref{tab:backpatching} presents the back-patching results. First we note the high success rates, correctly answering between 32\%-66\% of incorrect cases while keeping the correct cases at 100\% across models and token positions. This means that in many cases rerunning specific parts of the computation is enough in order to successfully extract the target information. On a more practical note, this also shows that significant gains can be made by correctly choosing the source and target layers, which we leave for future work. 

One can additionally observe that in LLaMA models back-patching $t_2$ achieves a higher success rate than back-patching $t_1$, which may point at the stages related to the last position of the prompt as being the more dominant failure point in answering two-hop queries. Moreover, this higher success rate is achieved despite the fact that back-patching $t_2$ does not assume any information about the structure of the prompt. We believe that this displays an additional strength of the method.

Figure \ref{fig:backpatching_e1_llama2-13b} displays the heat-map of the layers where back-patching $t_1$ succeeds for LLaMA 2 13B (see Appendix \ref{app:backpatching_heatmaps} for additional models). We first note the higher success rate when the source layers are the model's lower layers. These lower layers correlate to the resolving of the first hop, therefore back-patching them allows the model a better chance at solving the first hop. Second, we note the non existent success rate of the higher target layers. Unsurprisingly, in this case the model does not gain enough time to correctly complete the pathway, resulting in a low success rate.

Figure \ref{fig:backpatching_last_llama2-13b} displays the heat-map of the layers where back-patching $t_2$ succeeds for LLaMA 2 13B. Unlike when back-patching $t_1$, in this case the higher success rate can be found in the the upper source layers. These layers correlate to the resolving of the second hop at the last token position. Hence, back-patching a hidden representation taken from a later layer allows the model a better chance at solving the second hop. Furthermore, we note the higher success rate in lower target layers. This allows the model to gain time to resolve the second hop. This also gives the model access to parametric knowledge that might only be encoded in the lower layers, allowing the model to fully utilize its whole knowledge-base.

\section{Related Work}

Tracing and locating factual knowledge has been the focus of many recent works aiming to undercover and control the way transformers make predictions. Several of these works are motivated by knowledge editing \citep{de-cao-etal-2021-editing, mitchell2022fast, NEURIPS2022_6f1d43d5, zhang2024comprehensive} while others focus on understanding the inner workings of transformers \citep{dai-etal-2022-knowledge, geva-etal-2022-transformer, geva-etal-2023-dissecting}. Our work follows and expands on this second group by extending them to the more complex case of two-hop queries, which require multiple factual knowledge extractions and compositions.

The question whether models can implicitly reason over stored parametric knowledge has also gained recent popularity. One avenue of related research includes circuit discovery \citep{nanda2023progress, wang2023interpretability, NEURIPS2023_34e1dbe9, brinkmann-etal-2024-mechanistic}, but these works focus mainly on synthetic tasks. More closely related to this paper, recent works \citep{sakarvadia-etal-2023-memory, yang-etal-2024-large-language-models, li-etal-2024-understanding} have found existential evidence of latent reasoning in multi-hop queries, albeit without accounting for the flow of information throughout layers and positions. In a concurrent work by \citet{wang2024grokked} a multi-hop reasoning circuit similar to ours was discovered in a small transformer language model trained on synthetic data until grokking occurs. Our work finds further proof of such latent reasoning in a more practical setting that includes large language models, real world knowledge and a regular pretraining setup. We additionally use different mechanistic interpretability methods and account for the flow of information during the forward pass. 

Even without fully identifying the mechanism behind latent reasoning, several works have tried to address model's shortcomings in this area. One such work by \citet{sakarvadia-etal-2023-memory} assumes prior knowledge of the bridge entity, whose hidden representation is then strategically injected directly into the computation. Likewise, recent work by \citet{li-etal-2024-understanding} also assumes knowledge of the bridge entity and then employs knowledge editing techniques to improve performance, which is known to have potentially disruptive effects on models \citep{zhang2024comprehensive, hsueh2024editing, li2024unveiling}. Both these previously proposed methods have drawbacks, which a potential method based on back-patching would not share.
\section{Conclusion}

Our work uses mechanistic tools to study the latent reasoning abilities of LLMs on two hop queries. We observe strong evidence of a sequential latent reasoning pathway in which the second hop is answered in the mid-upper layers only after the first hop is answered by the lower layers. We find that this sequential nature could point to a possible limitation of transformers, where the second hop must be answered using only the knowledge encoded in the upper layers. To further validate this, we introduce and evaluate an analysis method named back-patching, based on extracting hidden representations from later layers and patching them back into earlier layers. Finally, we show that back-patching significantly improves performance on previously incorrect queries. Overall, our methods and findings open opportunities for understanding and improving latent reasoning in LLMs.
\section*{Limitations}

Several of our experiments rely on mechanistic methods that decode hidden representations and residual updates in various ways. While these methods have been widely used in many recent works \citep{geva-etal-2022-transformer, geva-etal-2023-dissecting, dar-etal-2023-analyzing, yang-etal-2024-large-language-models, li-etal-2024-understanding}, they can only be seen as an approximation. However, using multiple different techniques (such as both Patchscopes and sublayer projections) helps alleviate this concern. Additionally, our experiments are generally designed to give positive signals that result in lower bound results, hence we believe that these approximations do not undermine our findings.

In this work we study one prominent latent reasoning pathway, although others most likely exist \cite{mcgrath2023hydra, yang-etal-2024-large-language-models}. Additionally, we do not account for all possible parts of the discovered pathway (e.g., how the relations come in to play). Despite this, by focusing on the most relevant components we find sufficient proof for our main finding regarding the sequential nature of the computation.

Our work examines only two-hop queries, although we expect queries with three or more hops and even additional unrelated reasoning tasks to involve similar mechanisms to those we observed. This is due to previous \citep{wang2023interpretability, brinkmann-etal-2024-mechanistic} and concurrent \citep{wang2024grokked} works in different (if sometimes small or synthetic) settings observing pathways that can be broken down into concrete computational steps that match intuitive reasoning steps. These results resemble our finding that two-hop questions are broken down and answered sequentially by LLMs. We leave such expansion to additional settings for future work.

Despite the promising results of back-patching, our current proposed method is not a practical inference method, as only a subset of back-patches generate the correct answer. However, if one could efficiently select a subset of source and target layers a priori, this could make for a viable method. Although we believe that if achieving the best performance in multi-hop question answering is the sole goal, methods such as chain of thought prompting \citep{wei2022chain, press-etal-2023-measuring} would be far more effective.
\section*{Acknowledgements}

This work was supported by the Tel Aviv University Center for AI and Data Science (TAD) and the Israeli Science Foundation.

\bibliography{anthology,custom}

\appendix

\section{Technical Details}

We use the Baukit library \cite{bau2024} for several of the experiments. Each experiment was run on 1-4 A100 or H100 GPUs and lasted at most 24 hours. The models were retrieved and run using the HuggingFace Transformers library \cite{wolf-etal-2020-transformers}. When generating with sampling we use the default parameters from the Transformers library. We use half precision for the 70B model and full precision for all other models.

\section{Patchscopes First Decoded Layers} \label{app:ent_desc_decoded_layers}

\begin{figure*}[t]
\centering
\begin{subfigure}{.49\textwidth}
    \centering
    \includegraphics[scale=0.5]{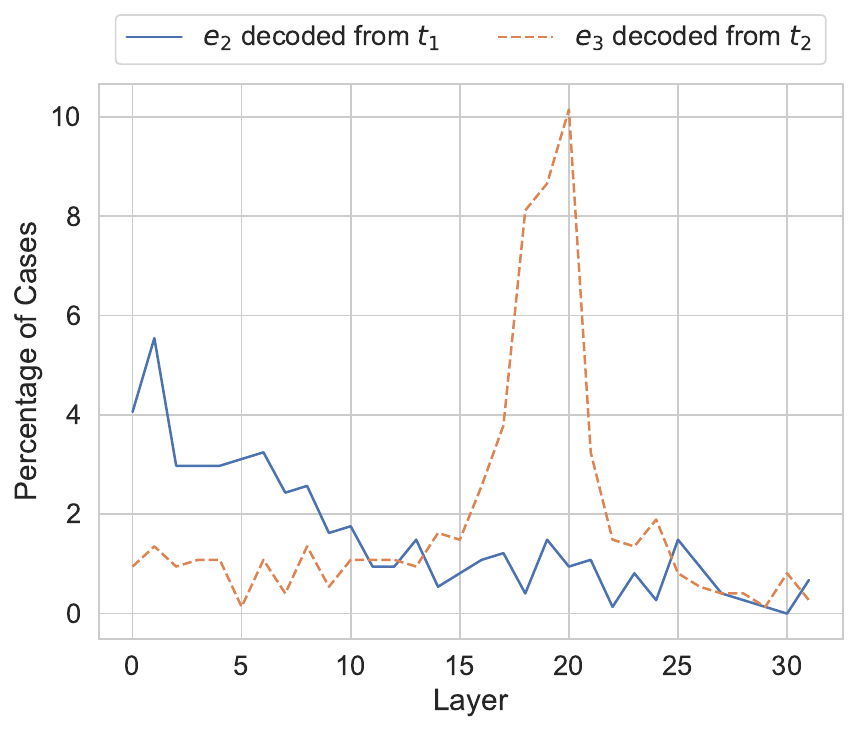}
    \caption{LLaMA 2 7B}    
    \label{fig:min_entity_description_layers_llama2-7b}
\end{subfigure}%
\hfill%
\begin{subfigure}{.49\textwidth}
    \centering
    \includegraphics[scale=0.5]{figures/min_entity_description_layers_llama2-13b.pdf}
    \caption{LLaMA 2 13B}    
    \label{fig:min_entity_description_layers_llama2-13b_apx}
\end{subfigure}
\par\bigskip
\begin{subfigure}{.49\textwidth}
    \centering
    \includegraphics[scale=0.5]{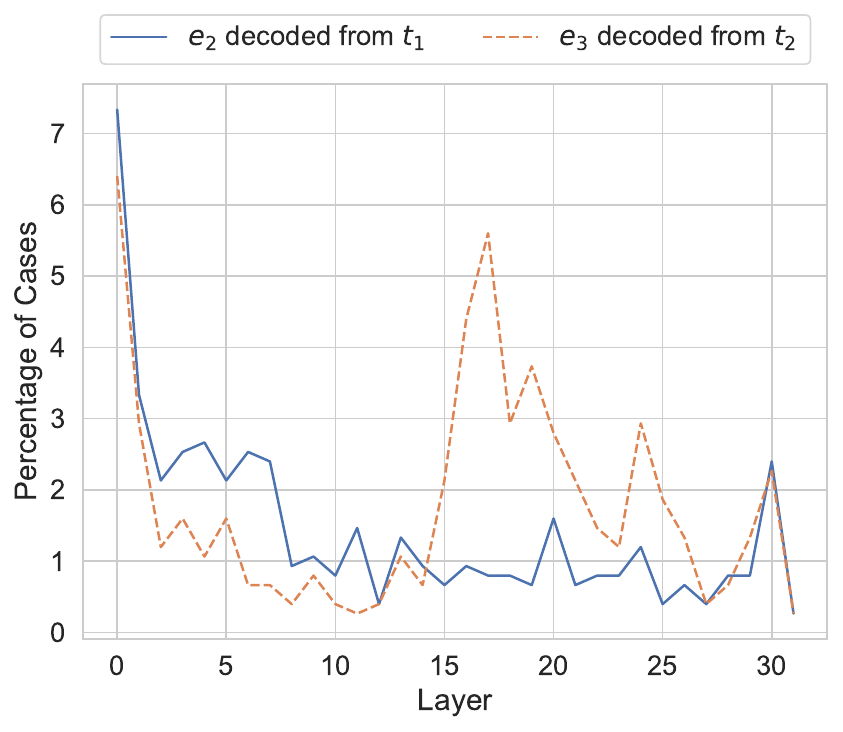}
    \caption{LLaMA 3 8B}    
    \label{fig:min_entity_description_layers_llama3-8b}
\end{subfigure}%
\hfill%
\begin{subfigure}{.49\textwidth}
    \centering
    \includegraphics[scale=0.5]{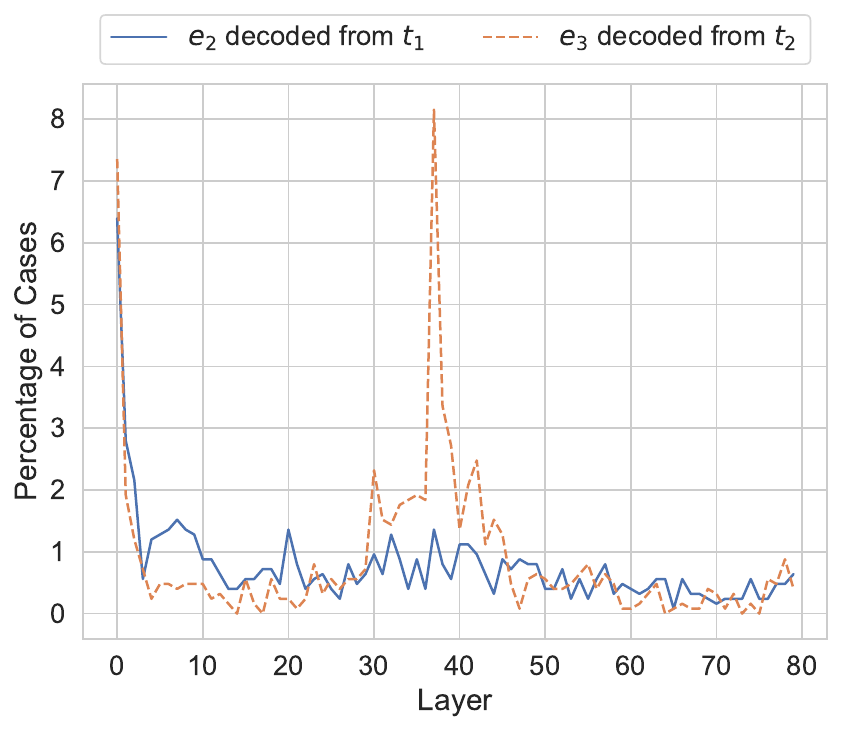}
    \caption{LLaMA 3 70B}    
    \label{fig:min_entity_description_layers_llama3-70b}
\end{subfigure}
\par\bigskip
\begin{subfigure}{.49\textwidth}
    \centering
    \includegraphics[scale=0.5]{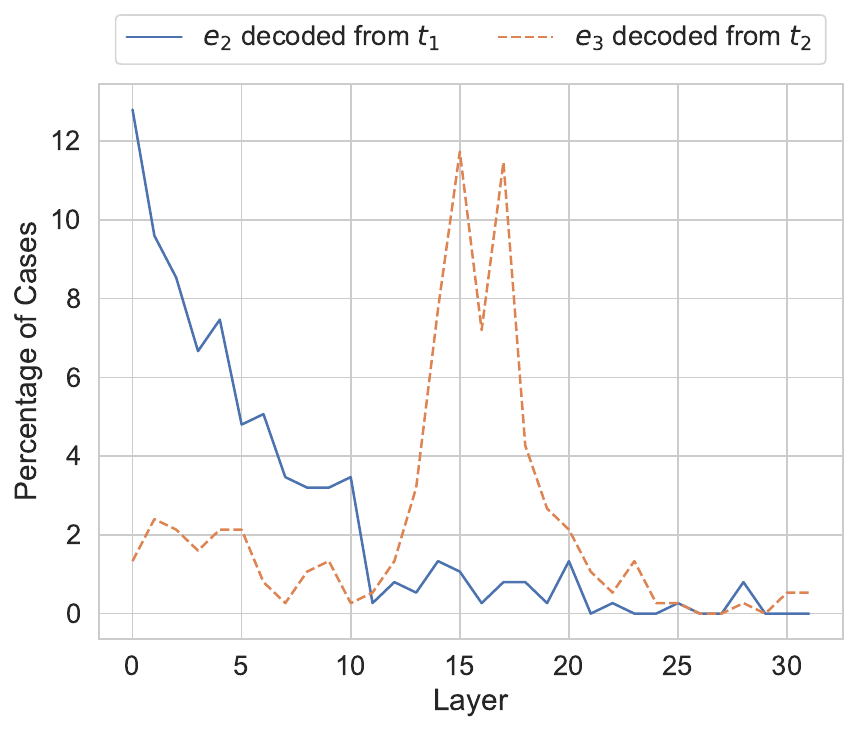}
    \caption{Pythia 6.9B}    
    \label{fig:min_entity_description_layers_pythia-6.9b}
\end{subfigure}%
\hfill%
\begin{subfigure}{.49\textwidth}
    \centering
    \includegraphics[scale=0.5]{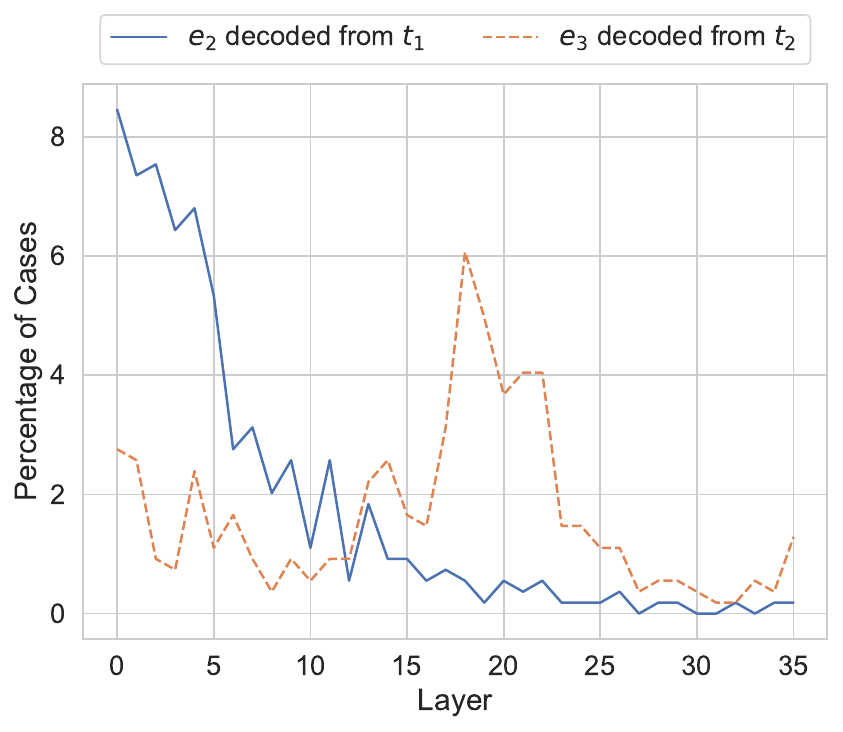}
    \caption{Pythia 12B}    
    \label{fig:min_entity_description_layers_pythia_12b}
\end{subfigure}
\caption[Patchscopes success by layer for all models]{Percentage of cases per layer where target entities were first successfully decoded using Patchscopes.}
\label{fig:min_entity_description_layers_all_models}
\end{figure*}

Figure \ref{fig:min_entity_description_layers_all_models} depicts the percentage of cases per layer where target entities were first successfully decoded using Patchscopes for all models (as described in \S\ref{sec:first_hop_resolved}).

\section{Patchscopes Heat-maps} \label{app:ent_desc_heatmaps}

\begin{figure*}[t]
\centering
\begin{subfigure}{.3\textwidth}
    \centering
    \includegraphics[scale=0.1]{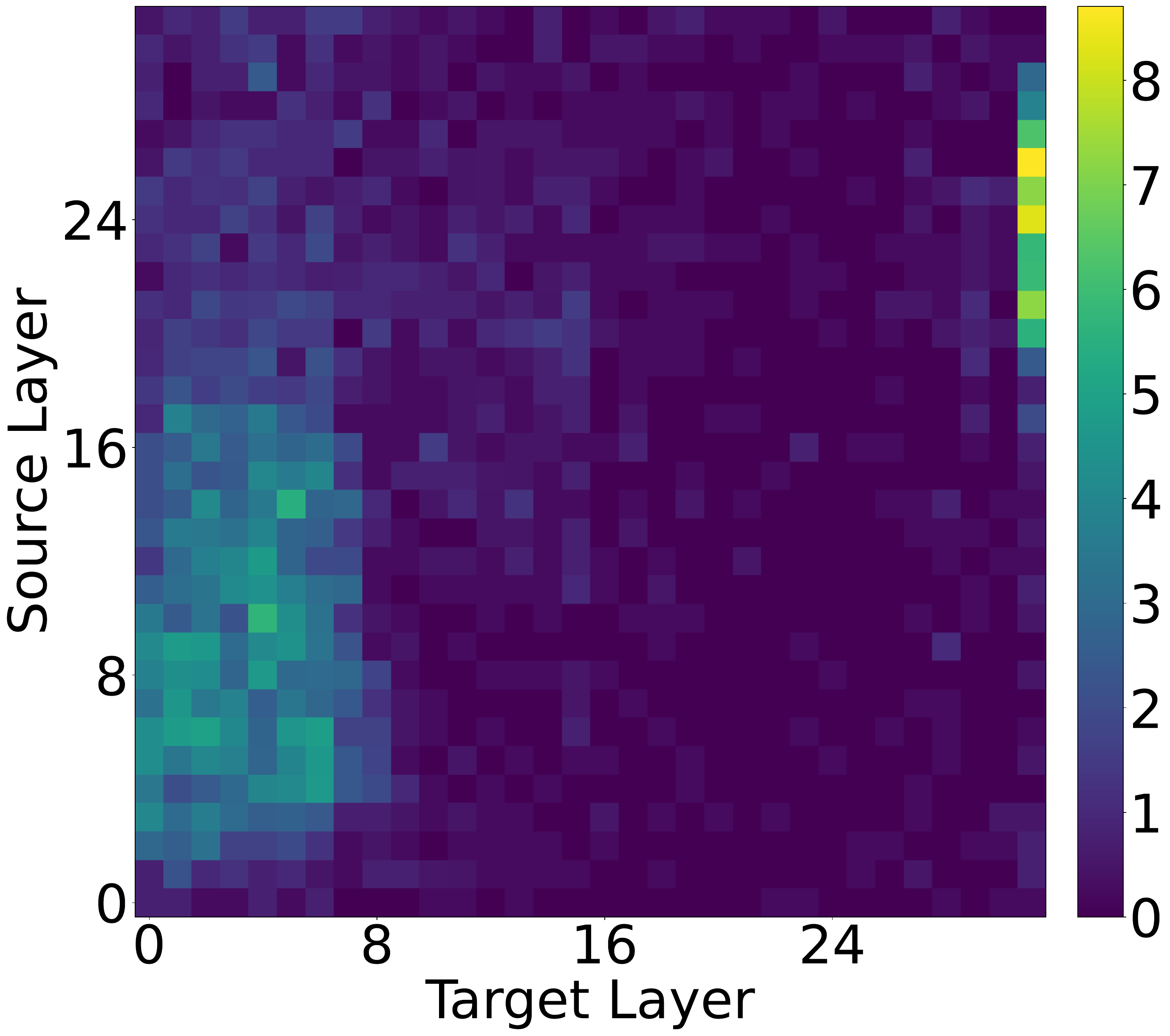}
    \caption{$e_2$ decoded from $t_1$, LLaMA 2 7B}
    \label{fig:entity_description_e1_e2_llama2-7b}
\end{subfigure}%
\hfill%
\begin{subfigure}{.3\textwidth}
    \centering
    \includegraphics[scale=0.1]{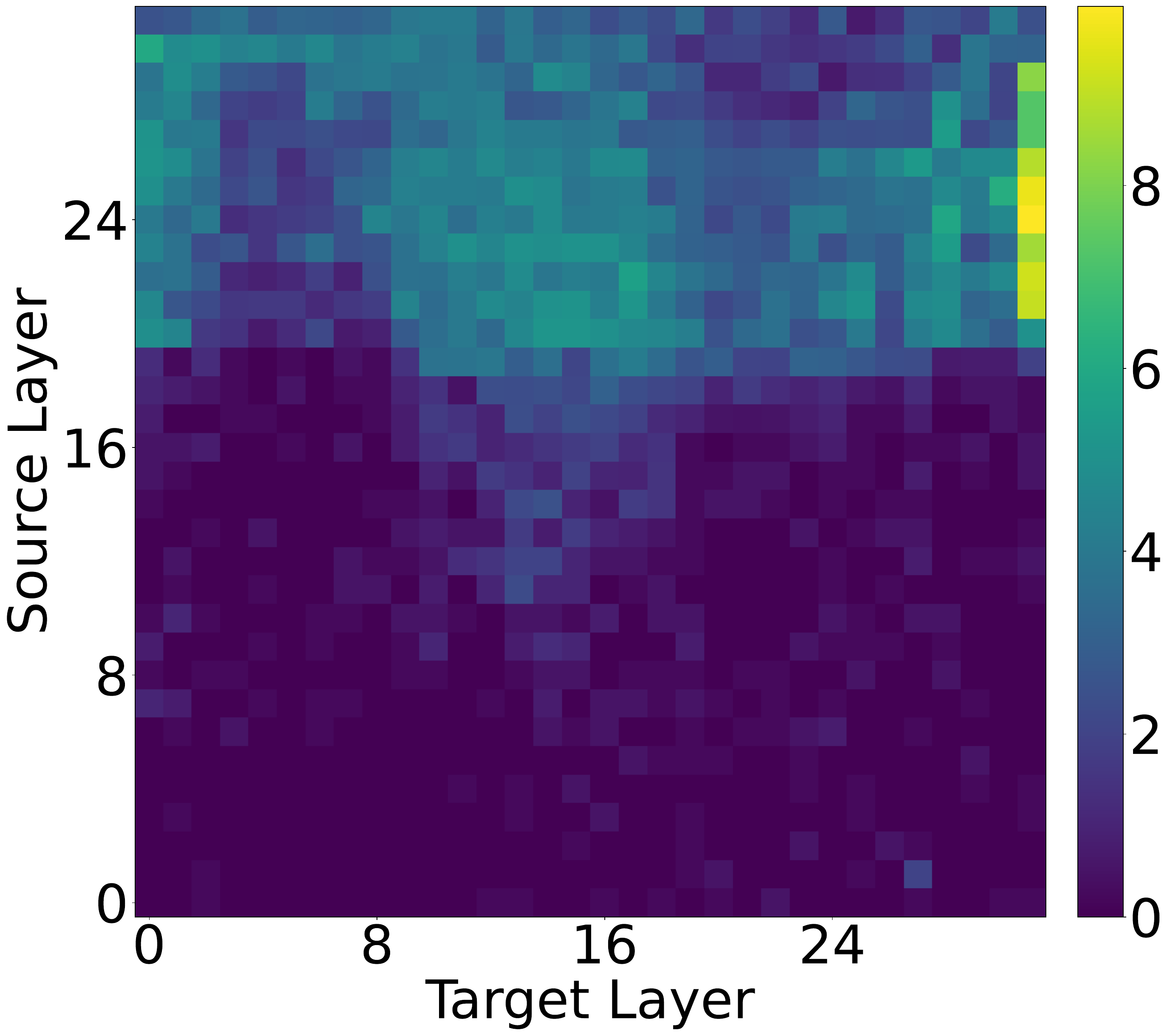}
    \caption{$e_2$ decoded from $t_2$, LLaMA 2 7B}
    \label{fig:entity_description_last_e2_llama2-7b}
\end{subfigure}%
\hfill%
\begin{subfigure}{.3\textwidth}
    \centering
    \includegraphics[scale=0.1]{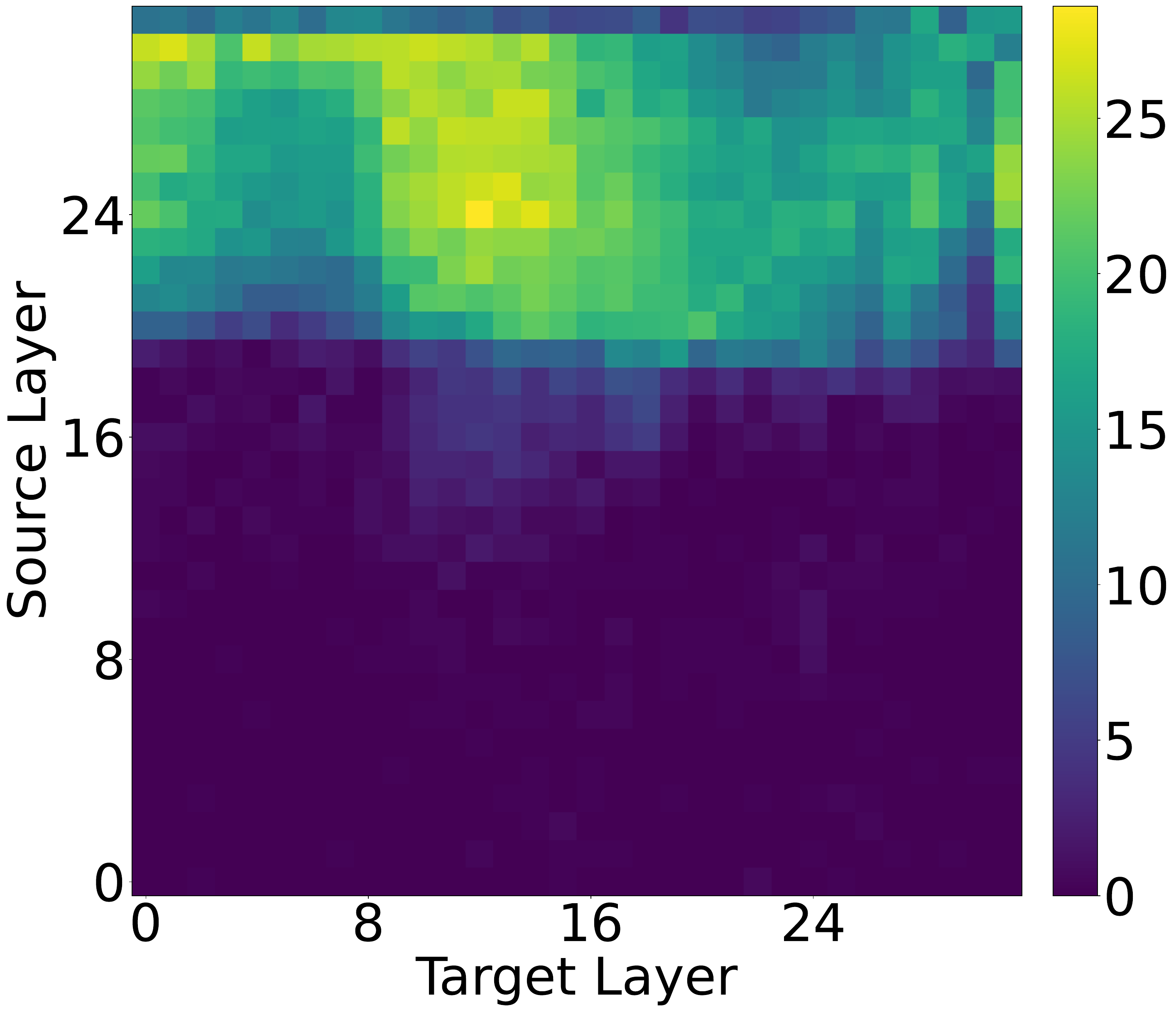}
    \caption{$e_3$ decoded from $t_2$, LLaMA 2 7B}
    \label{fig:entity_description_last_e3_llama2-7b}
\end{subfigure}
\par\bigskip
\begin{subfigure}{.3\textwidth}
    \centering
    \includegraphics[scale=0.1]{figures/entity_description_e1_e2_llama2-13b.pdf}
    \caption{$e_2$ decoded from $t_1$, LLaMA 2 13B}
    \label{fig:entity_description_e1_e2_llama2-13b_app}
\end{subfigure}%
\hfill%
\begin{subfigure}{.3\textwidth}
    \centering
    \includegraphics[scale=0.1]{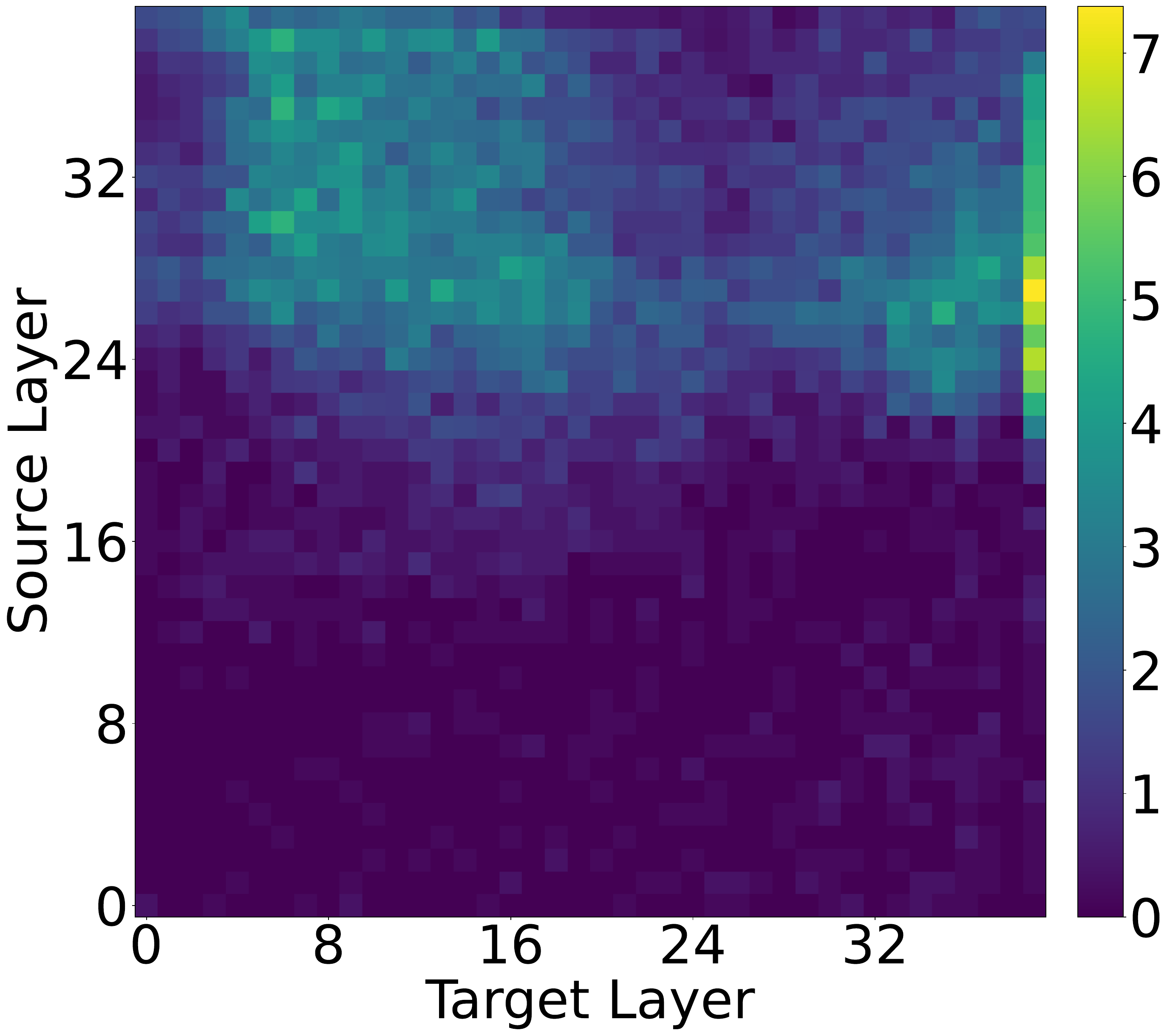}
    \caption{$e_2$ decoded from $t_2$, LLaMA 2 13B} 
    \label{fig:entity_description_last_e2_llama2-13b}
\end{subfigure}%
\hfill%
\begin{subfigure}{.3\textwidth}
    \centering
    \includegraphics[scale=0.1]{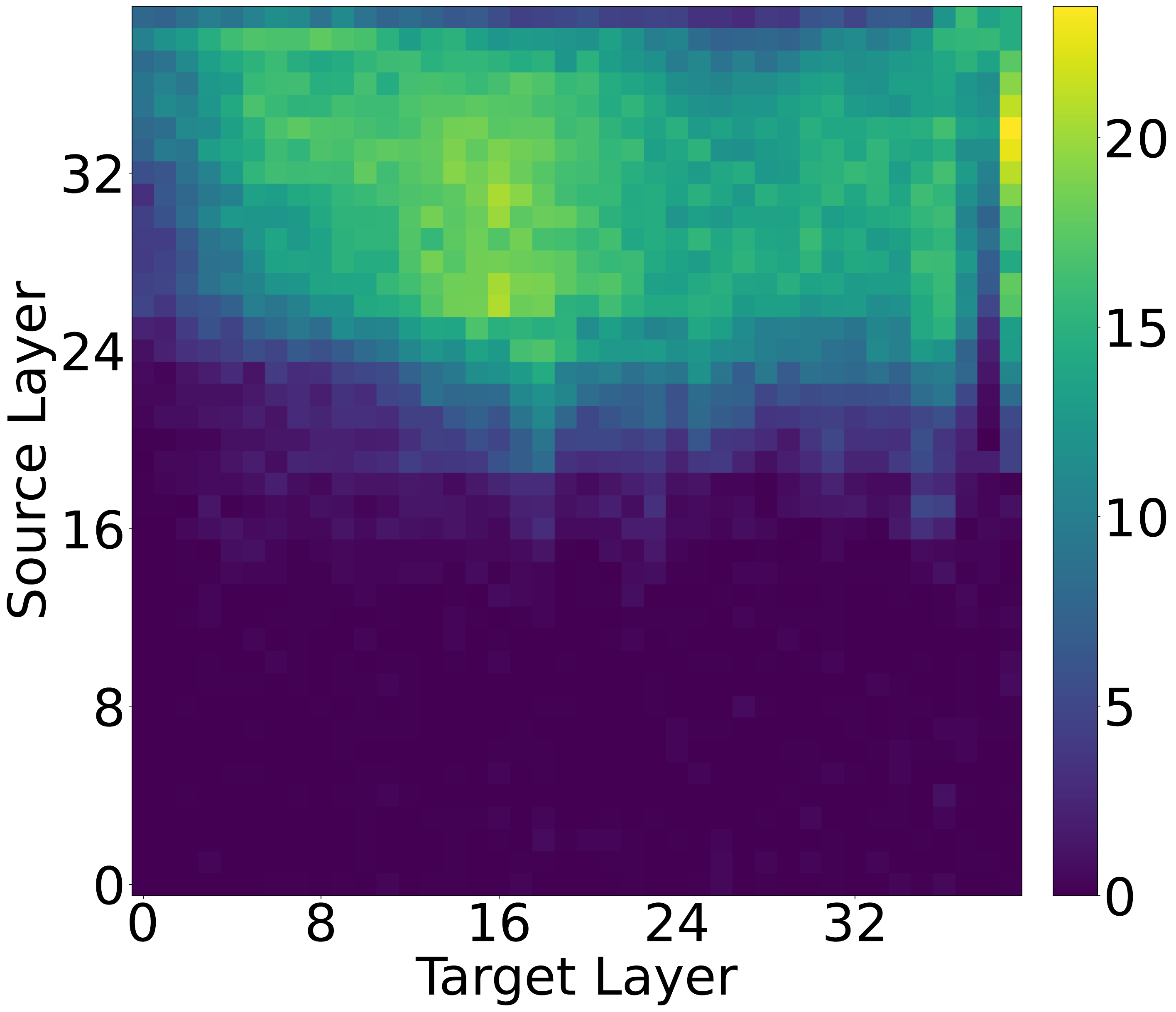}
    \caption{$e_3$ decoded from $t_2$, LLaMA 2 13B}
    \label{fig:entity_description_last_e3_llama2-13b}
\end{subfigure}
\par\bigskip
\begin{subfigure}{.3\textwidth}
    \centering
    \includegraphics[scale=0.1]{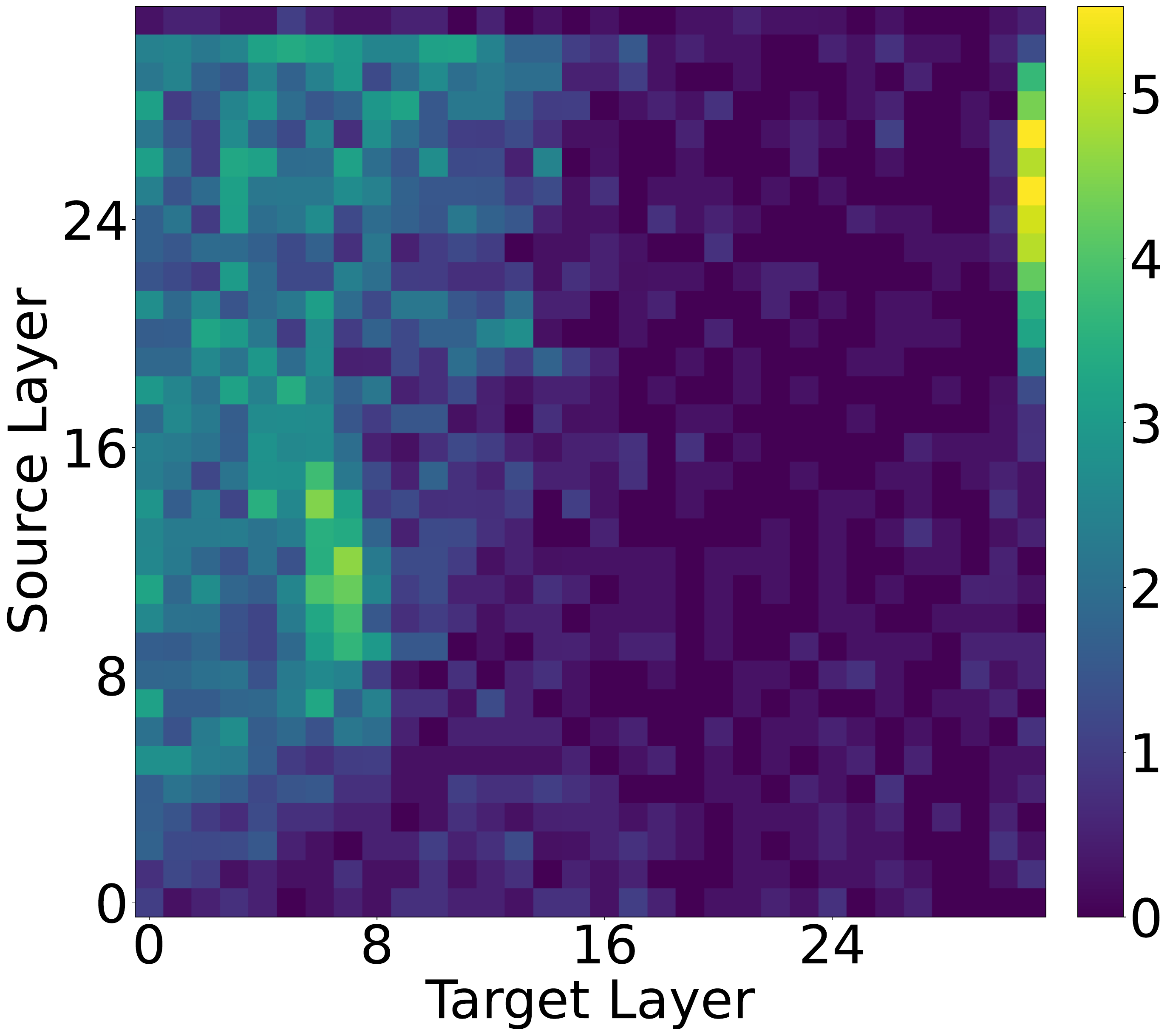}
    \caption{$e_2$ decoded from $t_1$, LLaMA 3 8B}
    \label{fig:entity_description_e1_e2_llama3-8b}
\end{subfigure}%
\hfill%
\begin{subfigure}{.3\textwidth}
    \centering
    \includegraphics[scale=0.1]{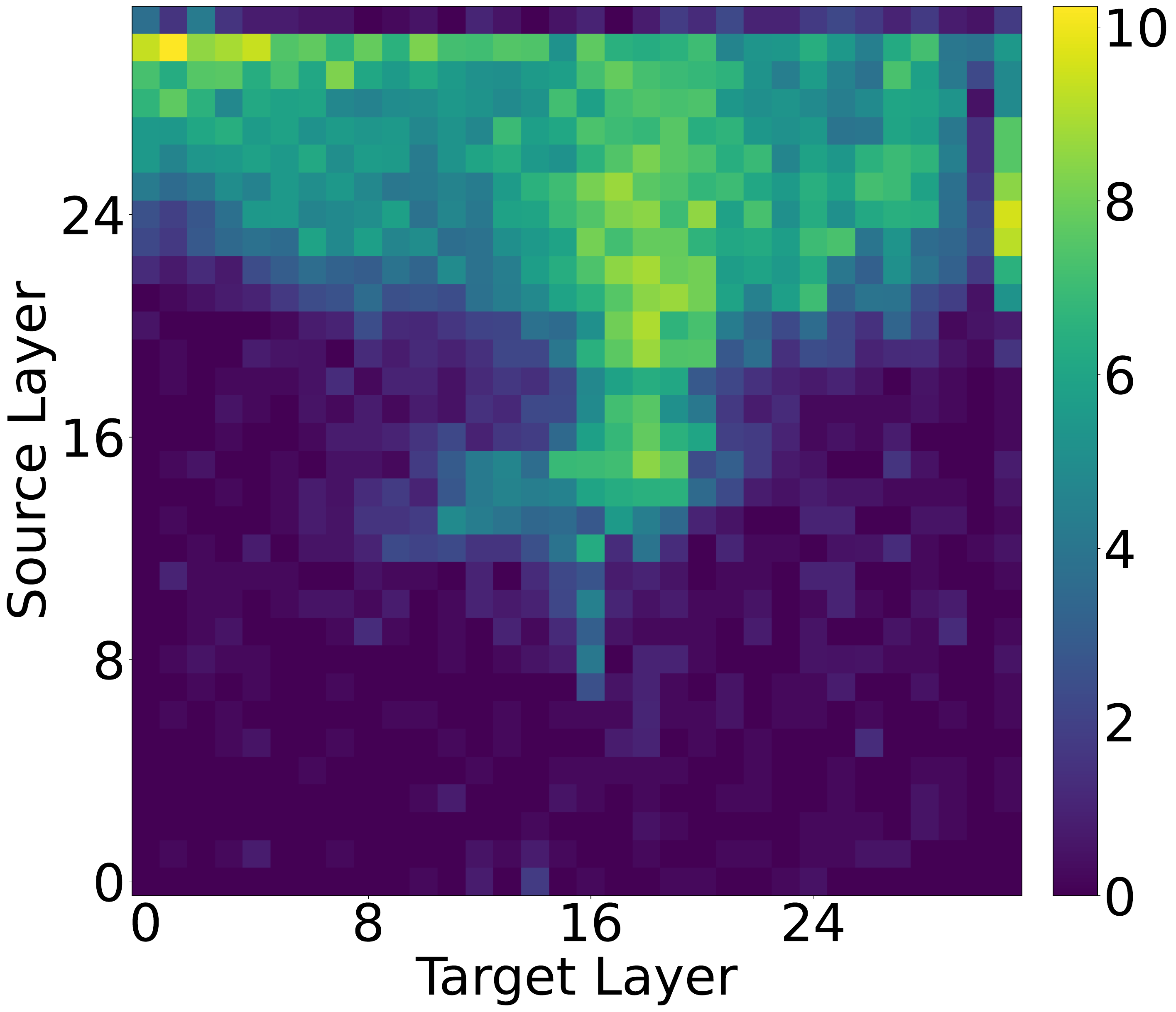}
    \caption{$e_2$ decoded from $t_2$, LLaMA 3 8B}    
    \label{fig:entity_description_last_e2_llama3-8b}
\end{subfigure}%
\hfill%
\begin{subfigure}{.3\textwidth}
    \centering
    \includegraphics[scale=0.1]{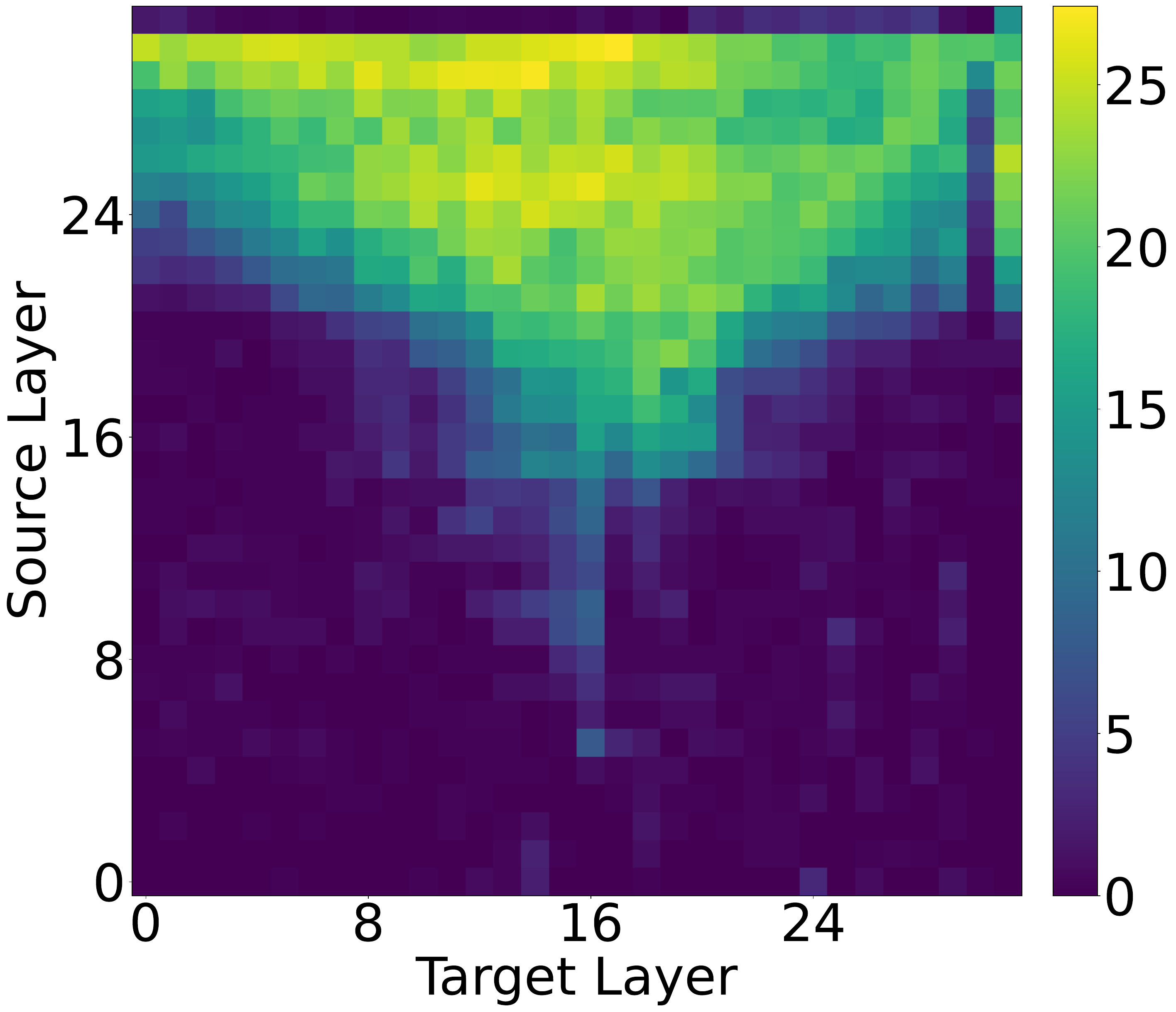}
    \caption{$e_3$ decoded from $t_2$, LLaMA 3 8B}    
    \label{fig:entity_description_last_e3_llama3-8b}
\end{subfigure}
\par\bigskip
\begin{subfigure}{.3\textwidth}
    \centering
    \includegraphics[scale=0.1]{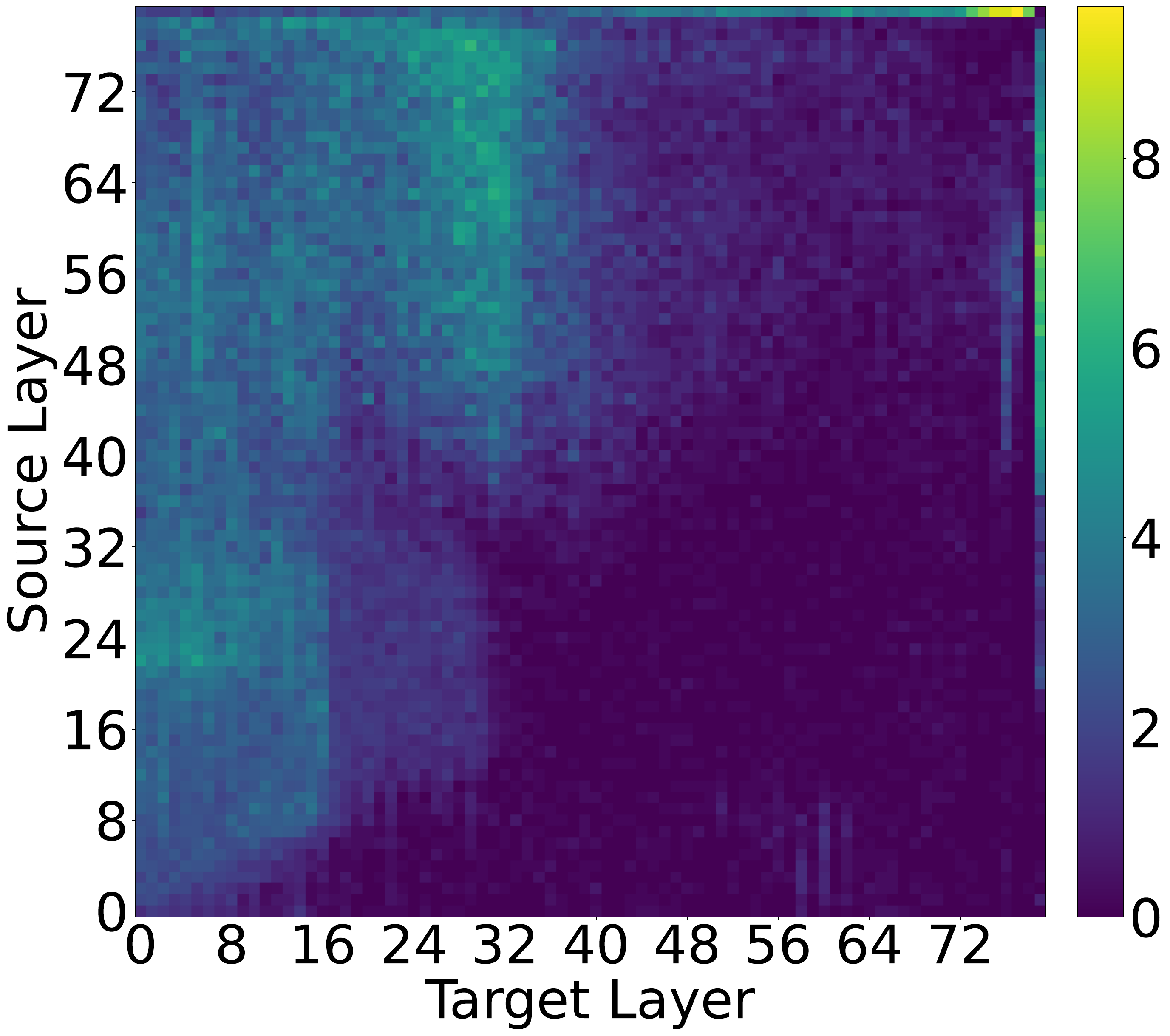}
    \caption{$e_2$ decoded from $t_1$, LLaMA 3 70B}
    \label{fig:entity_description_e1_e2_llama3-70b}
\end{subfigure}%
\hfill%
\begin{subfigure}{.3\textwidth}
    \centering
    \includegraphics[scale=0.1]{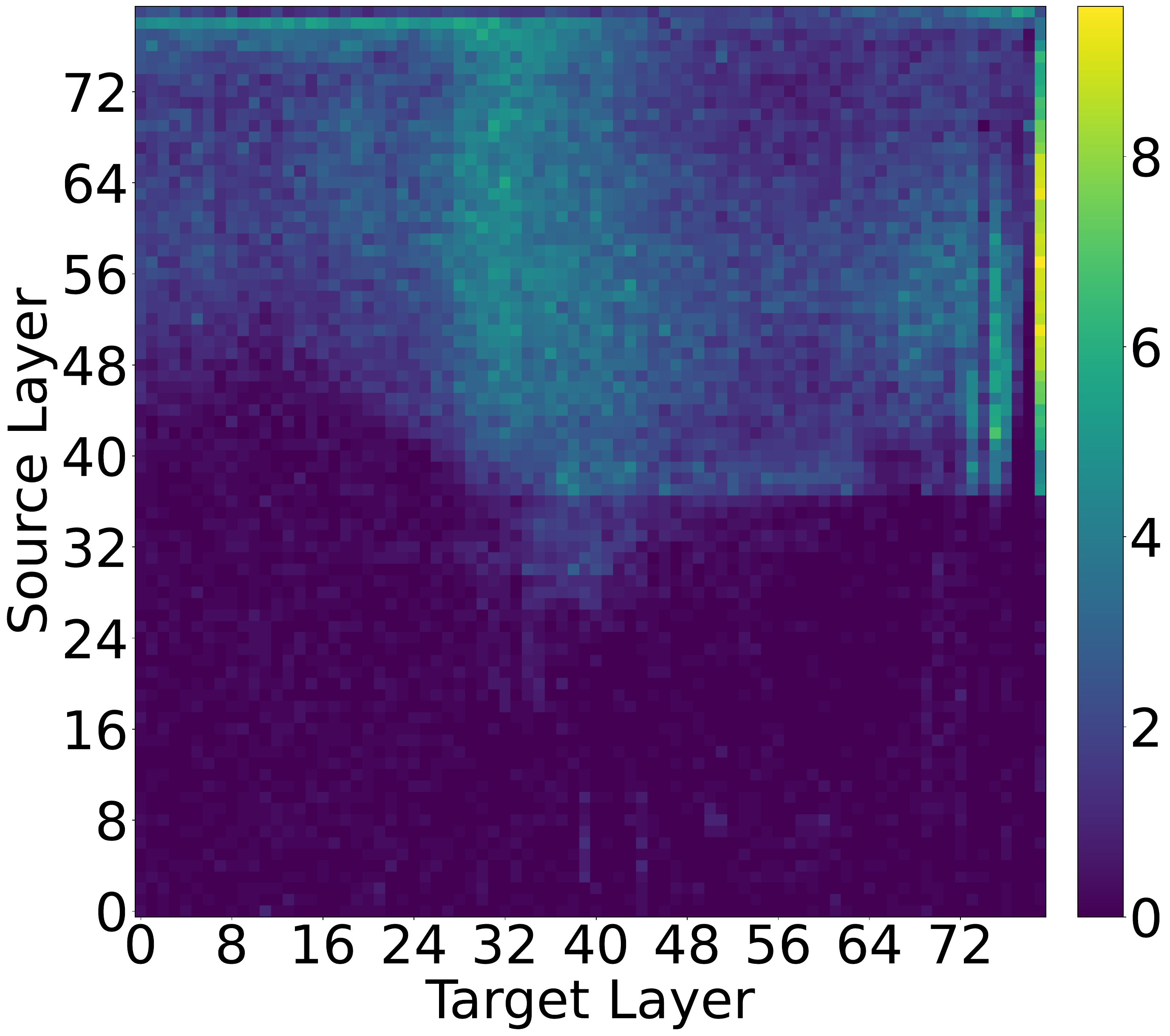}
    \caption{$e_2$ decoded from $t_2$, LLaMA 3 70B}    
    \label{fig:entity_description_last_e2_llama3-70b}
\end{subfigure}%
\hfill%
\begin{subfigure}{.3\textwidth}
    \centering
    \includegraphics[scale=0.1]{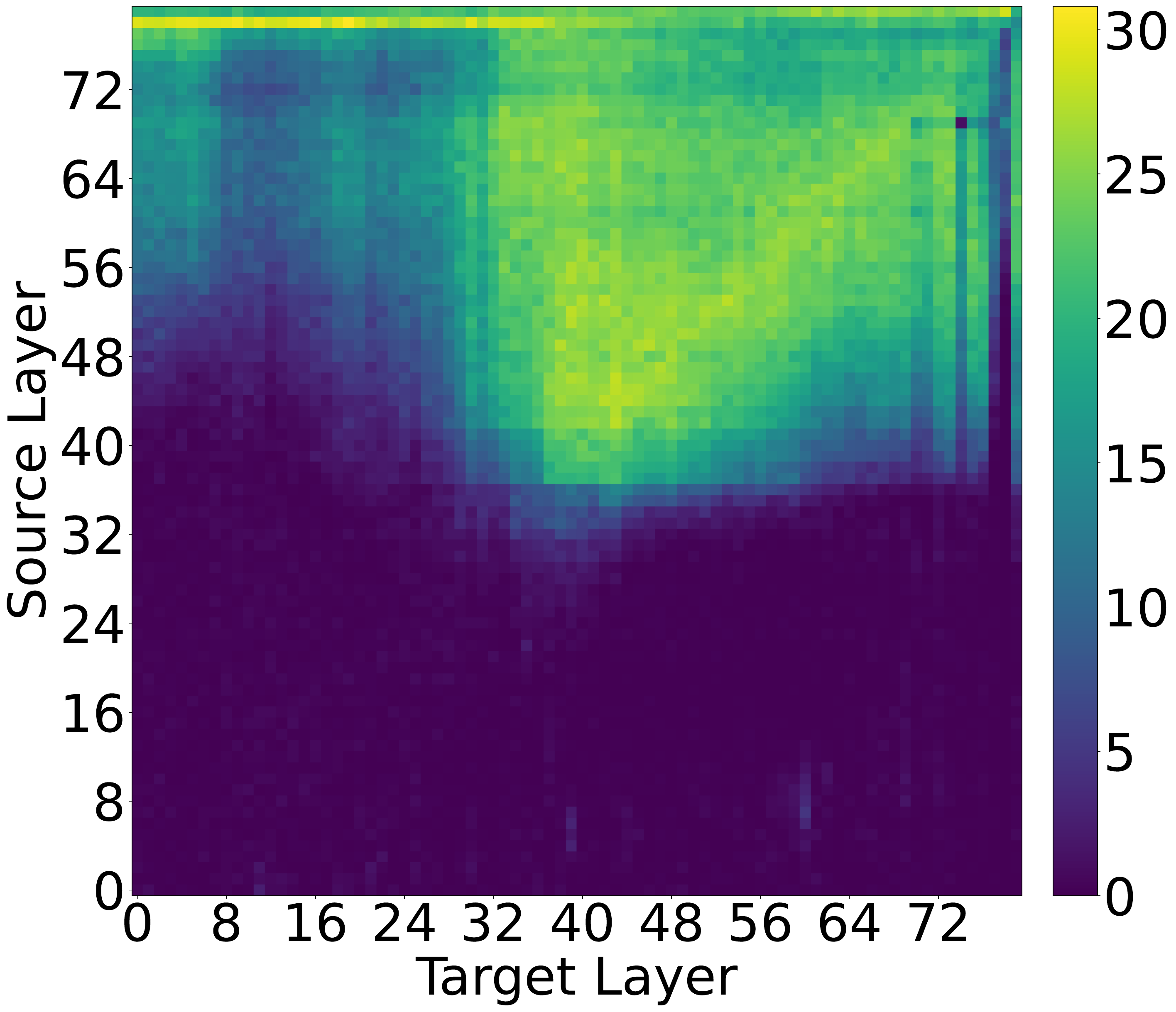}
    \caption{$e_3$ decoded from $t_2$, LLaMA 3 70B}    
    \label{fig:entity_description_last_e3_llama3-70b}
\end{subfigure}
\caption{Patchscopes success heat-maps for LLaMA models.}    
\label{fig:entity_description_llama_models}
\end{figure*}

\begin{figure*}[t]
\centering
\begin{subfigure}{.3\textwidth}
    \centering
    \includegraphics[scale=0.1]{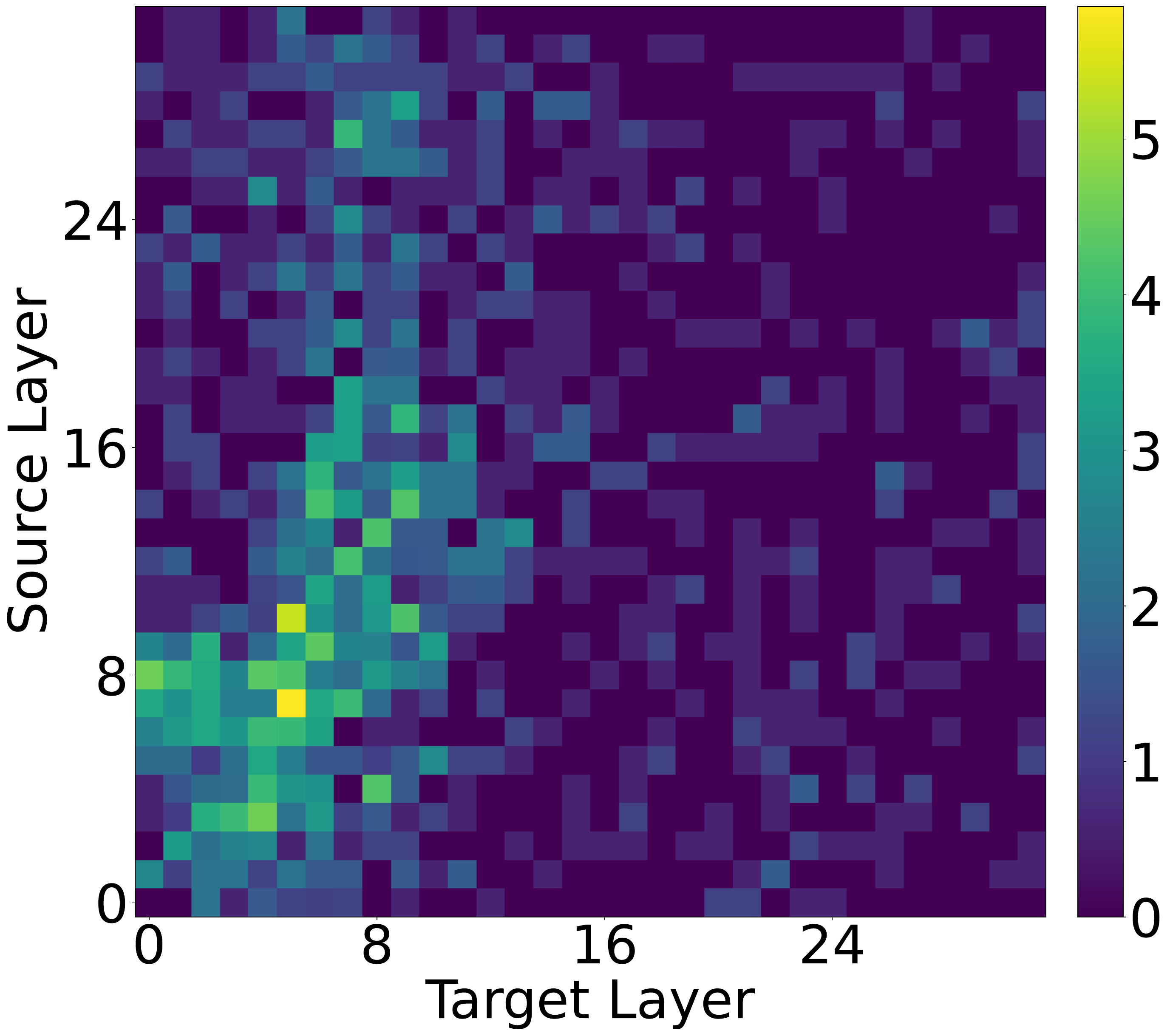}
    \caption{$e_2$ decoded from $t_1$, Pythia 6.9B}
    \label{fig:entity_description_e1_e2_pythia-6.9b}
\end{subfigure}%
\hfill%
\begin{subfigure}{.3\textwidth}
    \centering
    \includegraphics[scale=0.1]{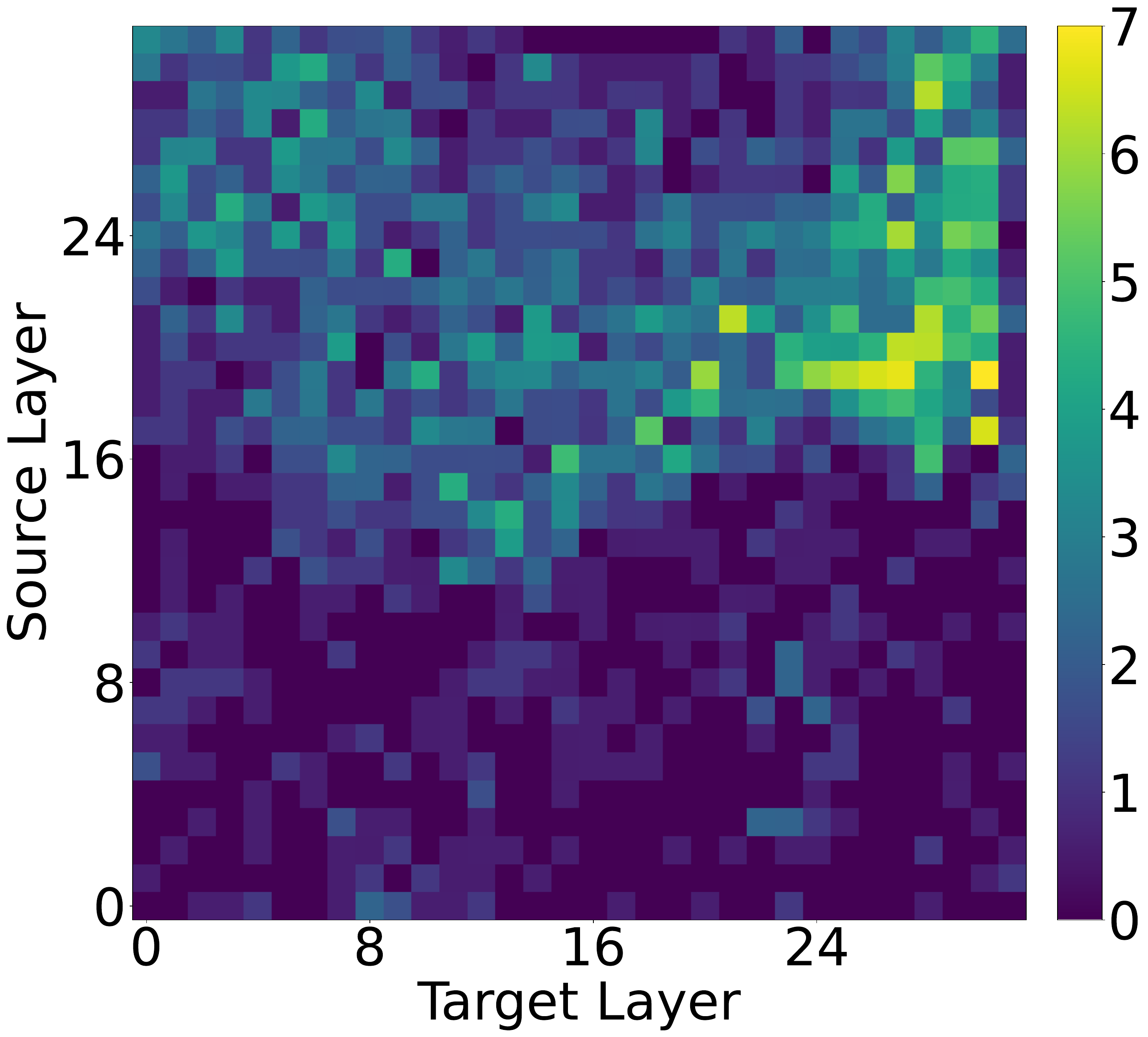}
    \caption{$e_2$ decoded from $t_2$, Pythia 6.9B}
    \label{fig:entity_description_last_e2_pythia-6.9b}
\end{subfigure}%
\hfill%
\begin{subfigure}{.3\textwidth}
    \centering
    \includegraphics[scale=0.1]{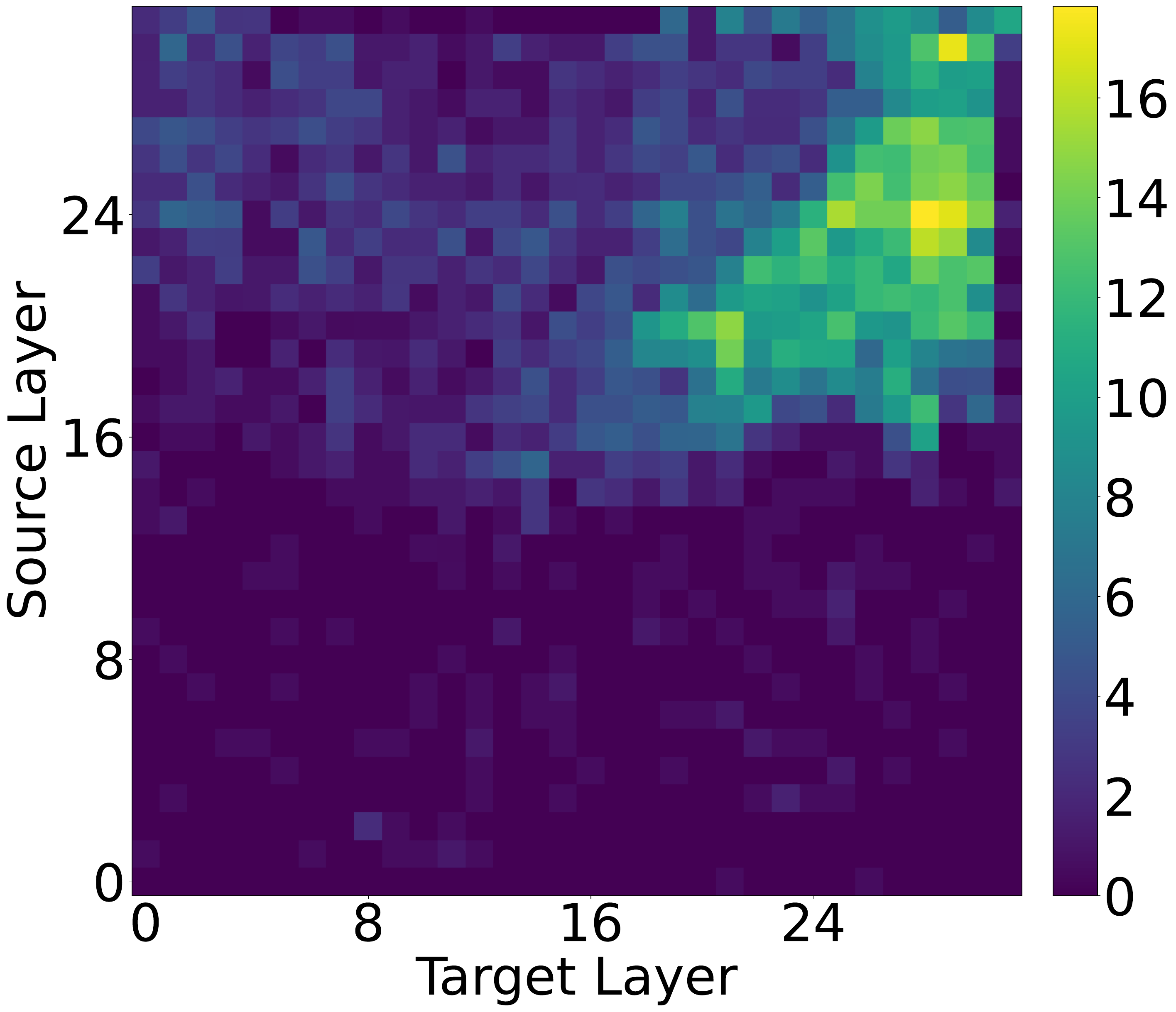}
    \caption{$e_3$ decoded from $t_2$, Pythia 6.9B}
    \label{fig:entity_description_last_e3_pythia-6.9b}
\end{subfigure}
\par\bigskip
\begin{subfigure}{.3\textwidth}
    \centering
    \includegraphics[scale=0.1]{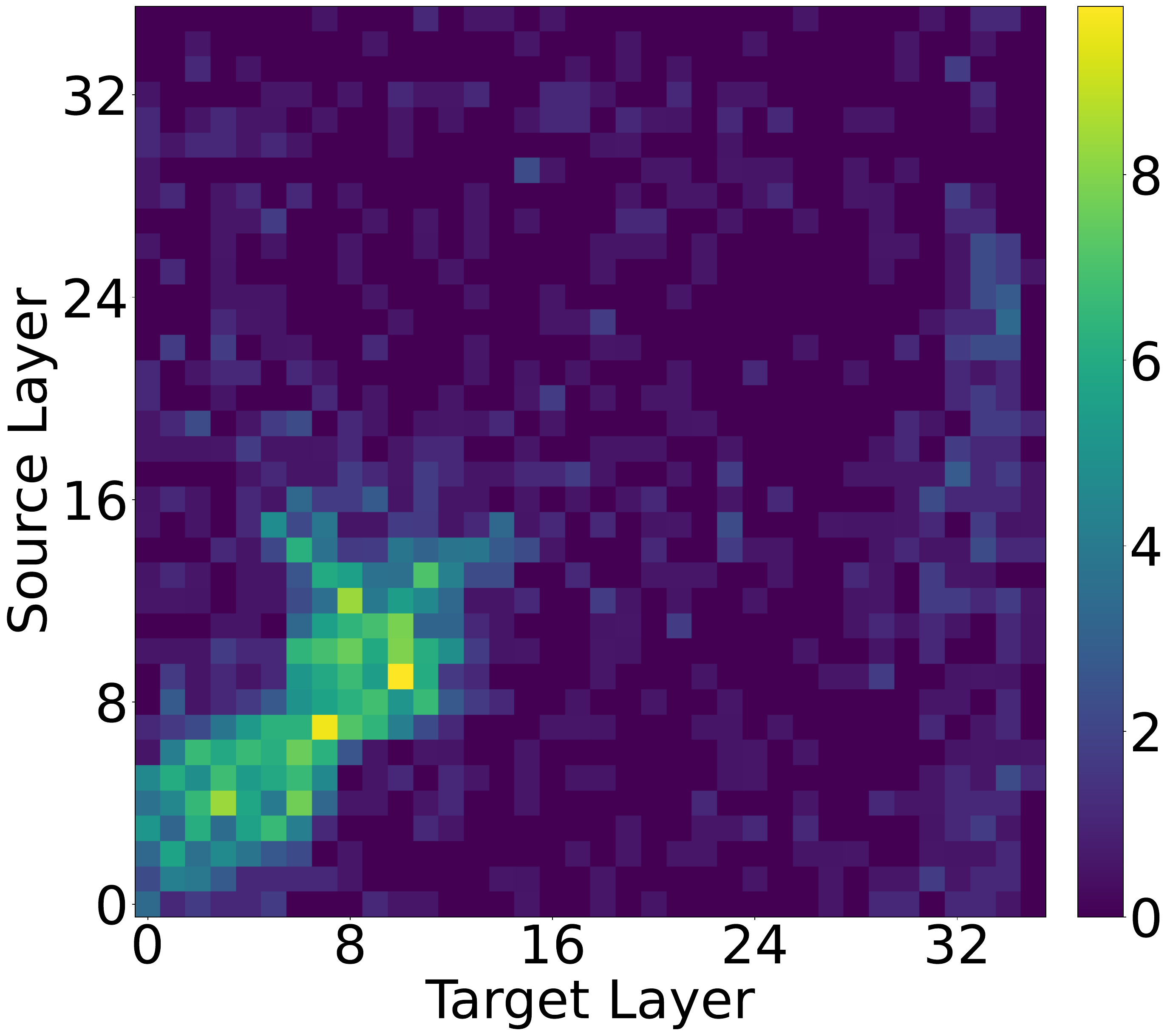}
    \caption{$e_2$ decoded from $t_1$, Pythia 12B}
    \label{fig:entity_description_e1_e2_pythia-12b}
\end{subfigure}%
\hfill%
\begin{subfigure}{.3\textwidth}
    \centering
    \includegraphics[scale=0.1]{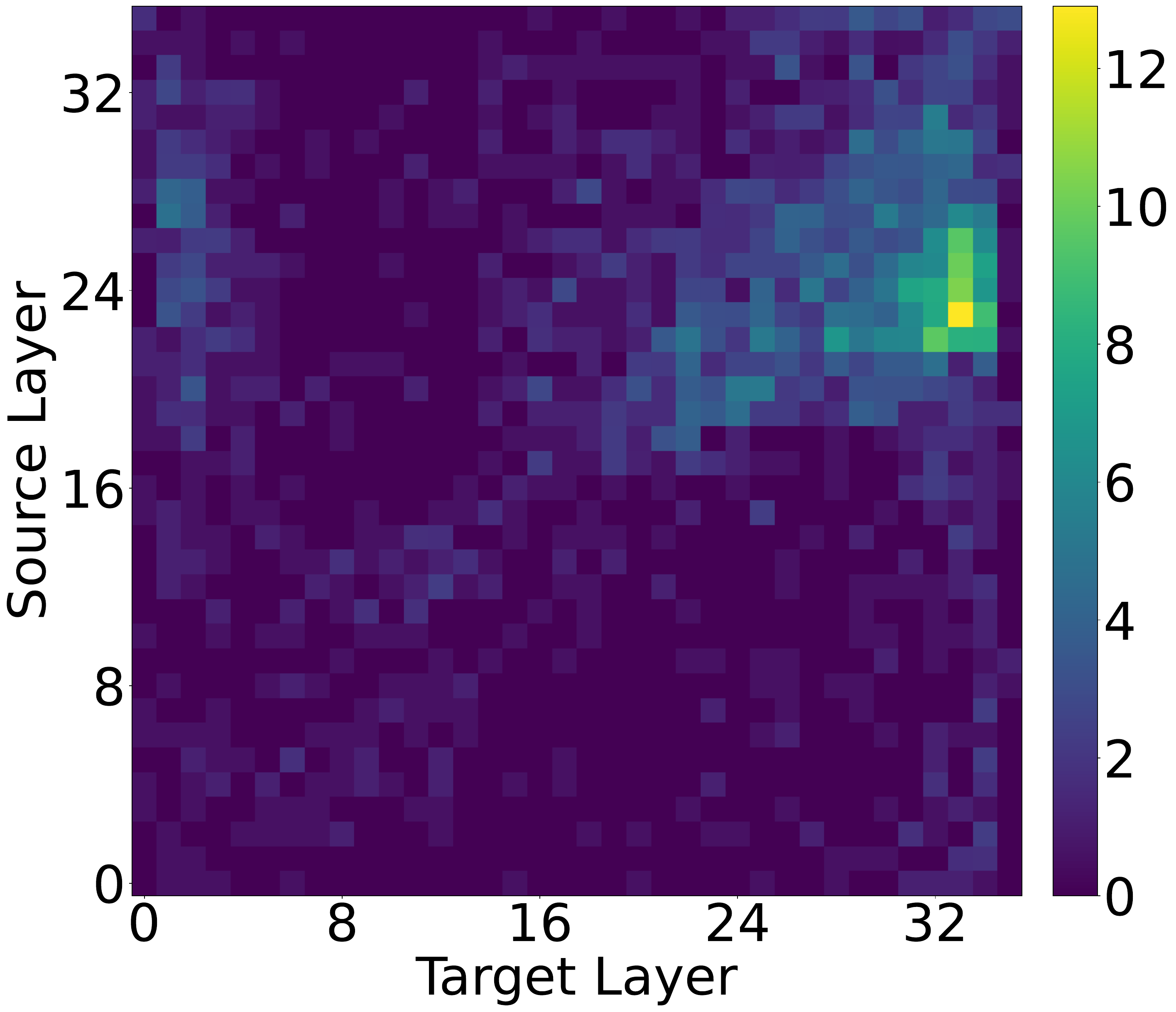}
    \caption{$e_2$ decoded from $t_2$, Pythia 12B} 
    \label{fig:entity_description_last_e2_pythia-12b}
\end{subfigure}%
\hfill%
\begin{subfigure}{.3\textwidth}
    \centering
    \includegraphics[scale=0.1]{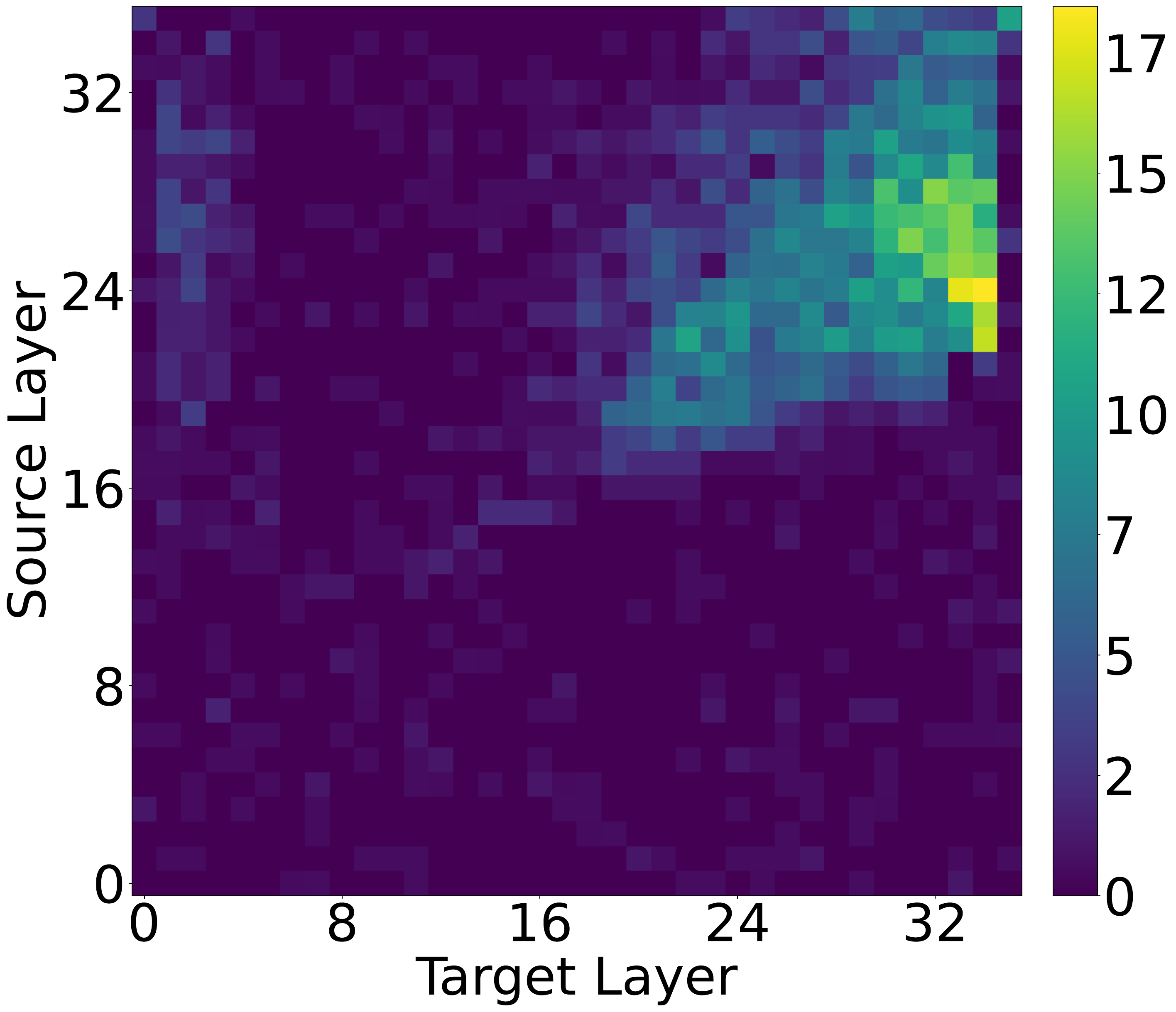}
    \caption{$e_3$ decoded from $t_2$, Pythia 12B}
    \label{fig:entity_description_last_e3_pythia-12b}
\end{subfigure}
\caption{Patchscopes success heat-maps for Pythia models.}    
\label{fig:entity_description_pythia_models}
\end{figure*}

Figures \ref{fig:entity_description_llama_models} and \ref{fig:entity_description_pythia_models} depict the heat-maps of Patchscope success cases as introduced in \S\ref{sec:first_hop_resolved}. We report Patchscope experiments for decoding $e_2$ from $t_1$, $e_2$ from $t_2$, and $e_3$ from $t_2$ for all models.

\section{First Observed Layers of Pathway Stages} \label{app:layers_of_stages}

\begin{figure*}[t]
\centering
\begin{subfigure}{.5\textwidth}
    \centering
    \includegraphics[scale=0.47]{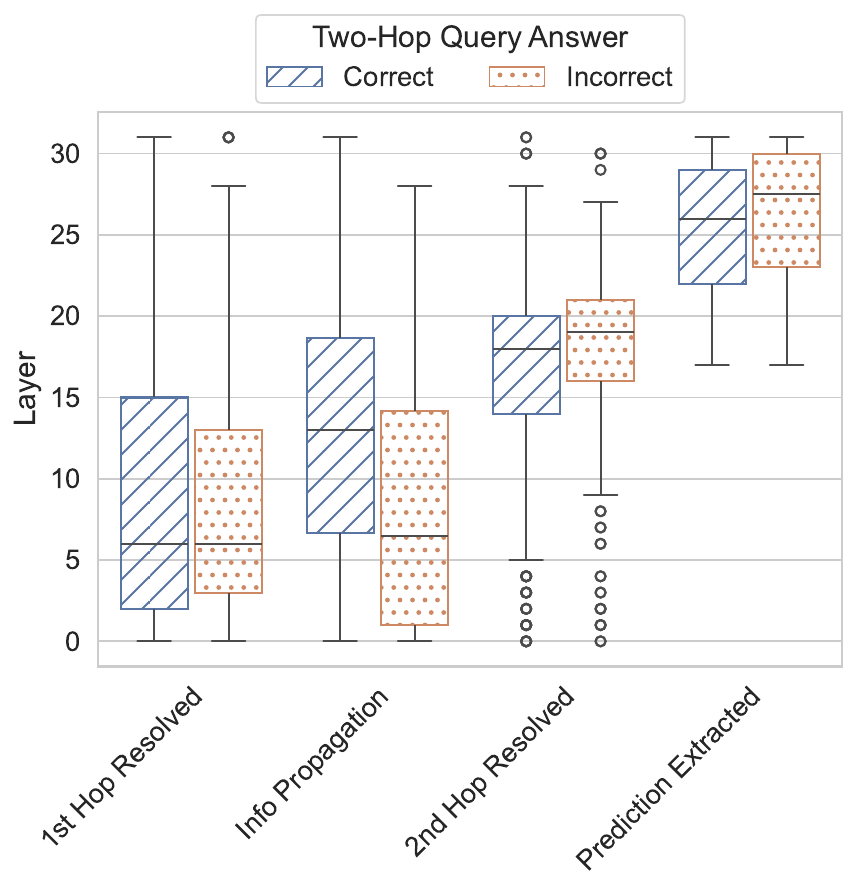}
    \caption{LLaMA 2 7B}    
    \label{fig:stages_boxplot_llama2-7b}
\end{subfigure}%
\hfill%
\begin{subfigure}{.5\textwidth}
    \centering
    \includegraphics[scale=0.47]{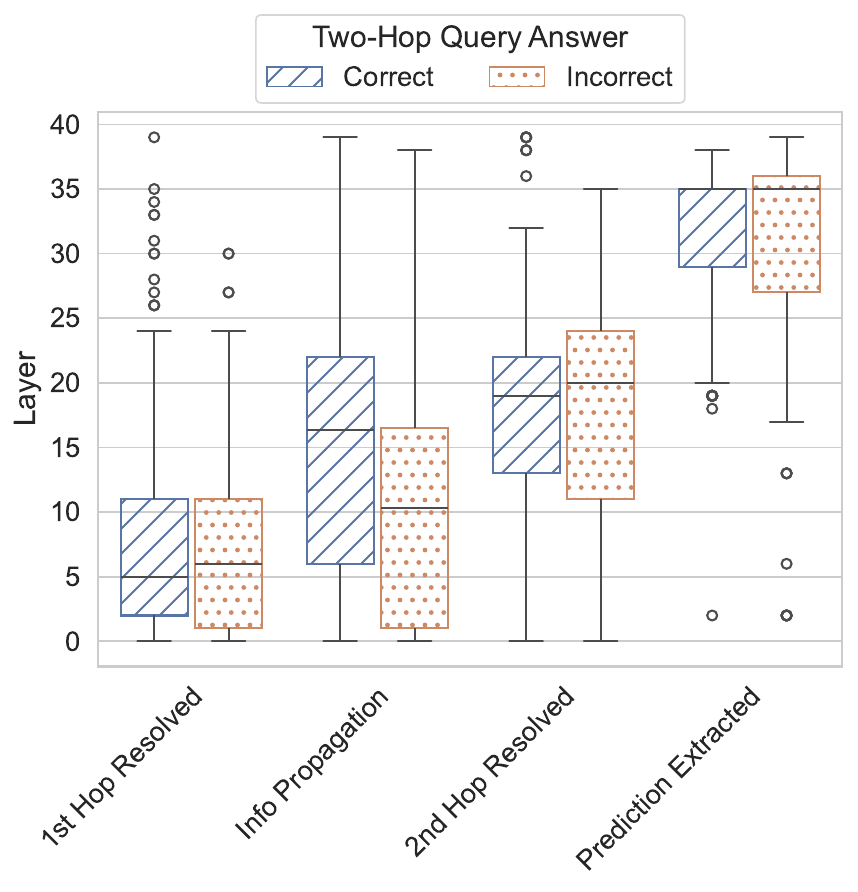}
    \caption{LLaMA 2 13B}    
    \label{fig:stages_boxplot_llama2-13b}
\end{subfigure}
\par\bigskip
\begin{subfigure}{.5\textwidth}
    \centering
    \includegraphics[scale=0.47]{figures/stages_boxplot_llama3-8b.pdf}
    \caption{LLaMA 3 8B}    
    \label{fig:stages_boxplot_llama3-8b_app}
\end{subfigure}%
\hfill%
\begin{subfigure}{.5\textwidth}
    \centering
    \includegraphics[scale=0.47]{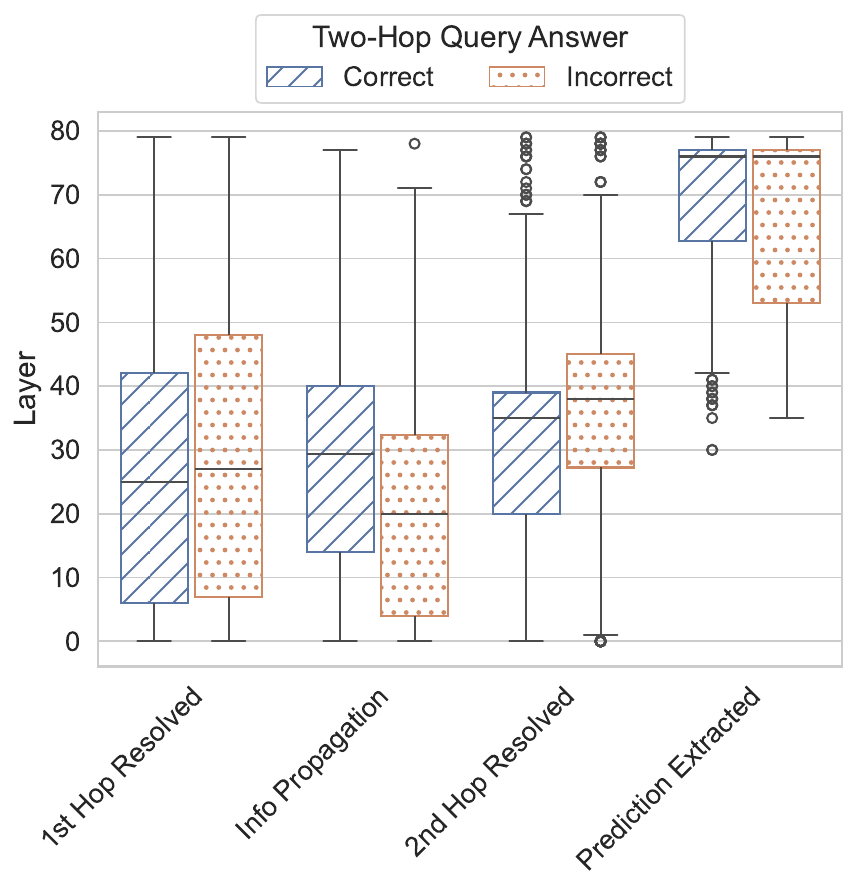}
    \caption{LLaMA 3 70B}    
    \label{fig:stages_boxplot_llama3-70b}
\end{subfigure}
\par\bigskip
\begin{subfigure}{.5\textwidth}
    \centering
    \includegraphics[scale=0.47]{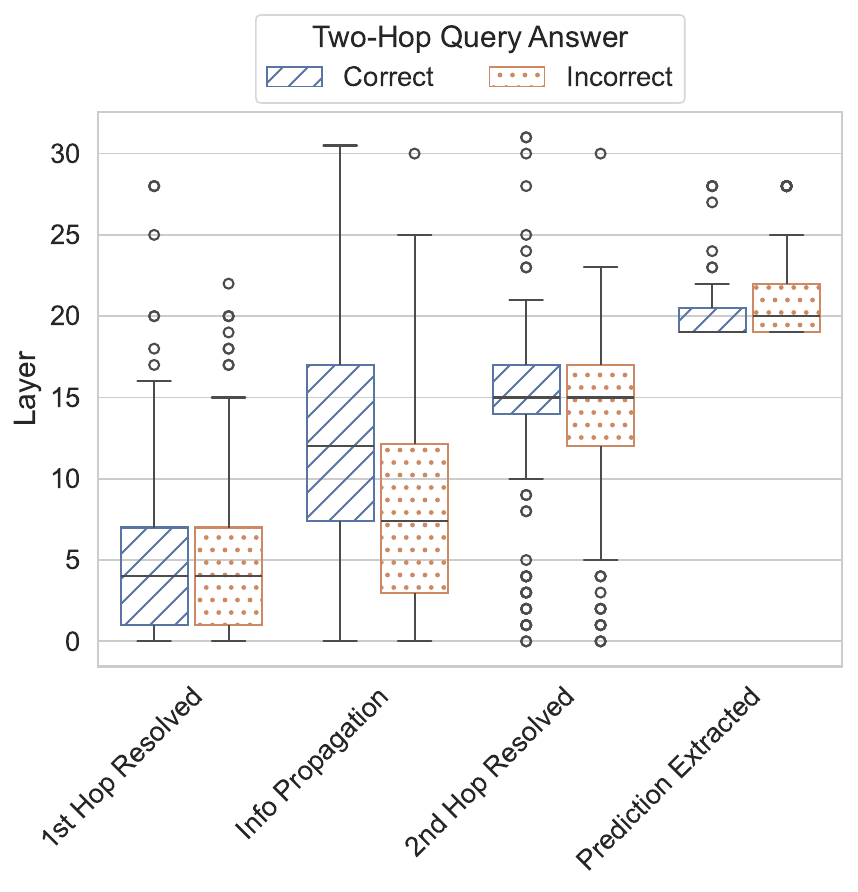}
    \caption{Pythia 6.9B}    
    \label{fig:stages_boxplot_pythia-6.9b}
\end{subfigure}%
\hfill%
\begin{subfigure}{.5\textwidth}
    \centering
    \includegraphics[scale=0.47]{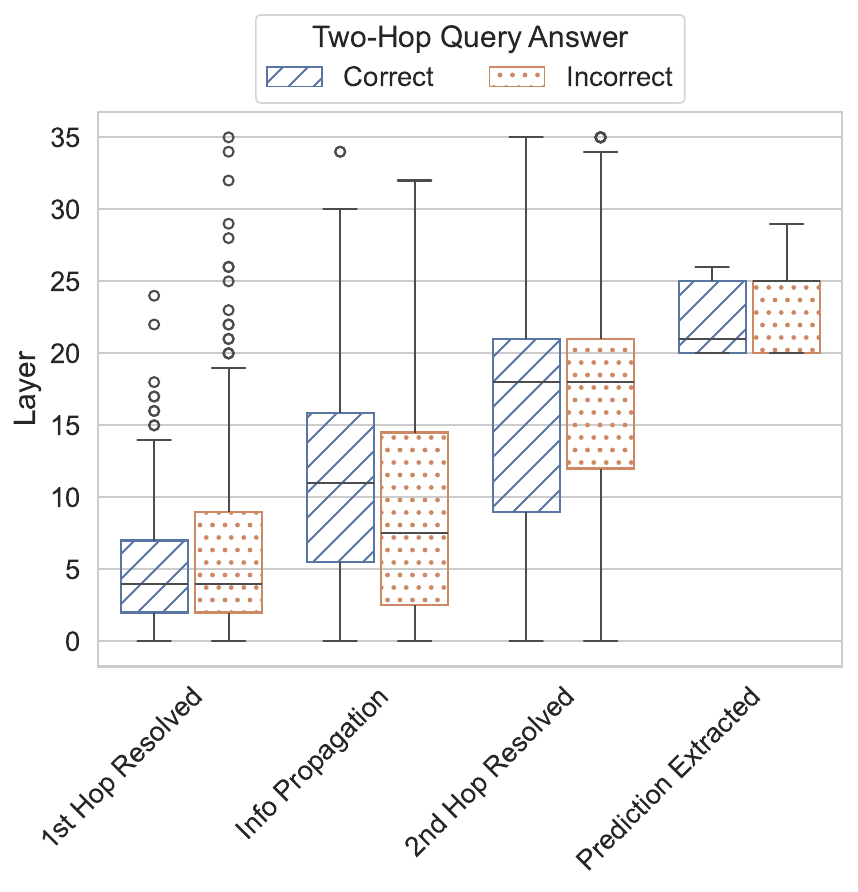}
    \caption{Pythia 12B}    
    \label{fig:stages_boxplot_pythia-12b}
\end{subfigure}
\caption{A comparison of the first layers of each pathway stage between correct and incorrect cases for all models.}    
\label{fig:stages_boxplot_all_models}
\end{figure*}

Figure \ref{fig:stages_boxplot_all_models} depicts a comparison of the first layers of each stage in the observed pathway between correct and incorrect cases (as shown in \S\ref{sec:info_propagates}) for all models.

\section{Back-patching Heat-maps} \label{app:backpatching_heatmaps}

\begin{figure*}[t]
\centering
\begin{subfigure}{.5\textwidth}
    \centering
    \includegraphics[scale=0.12]{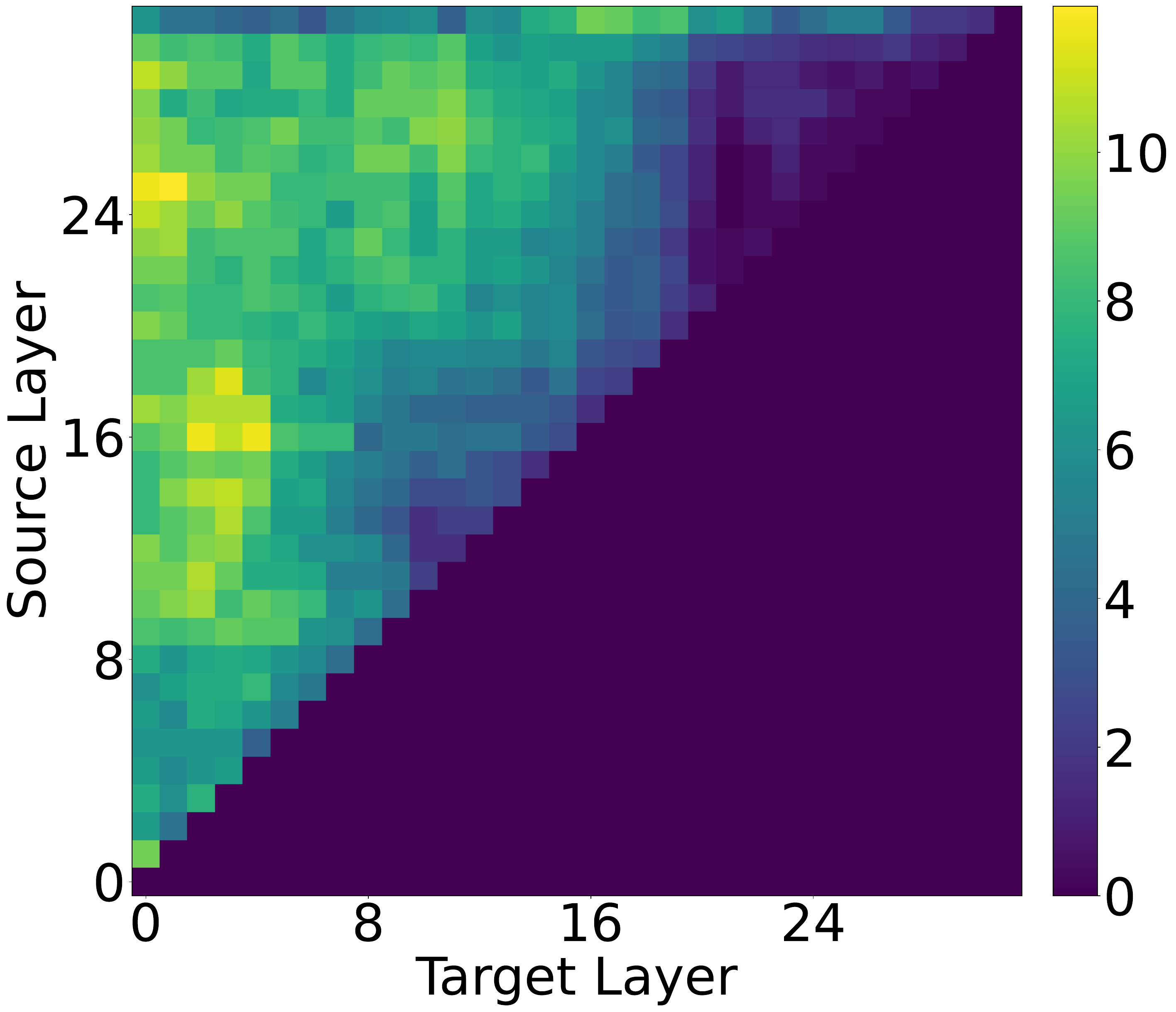}
    \caption{Source token is $t_1$, LLaMA 2 7B}    
    \label{fig:backpatching_e1_llama2-7b}
\end{subfigure}%
\hfill%
\begin{subfigure}{.5\textwidth}
    \centering
    \includegraphics[scale=0.12]{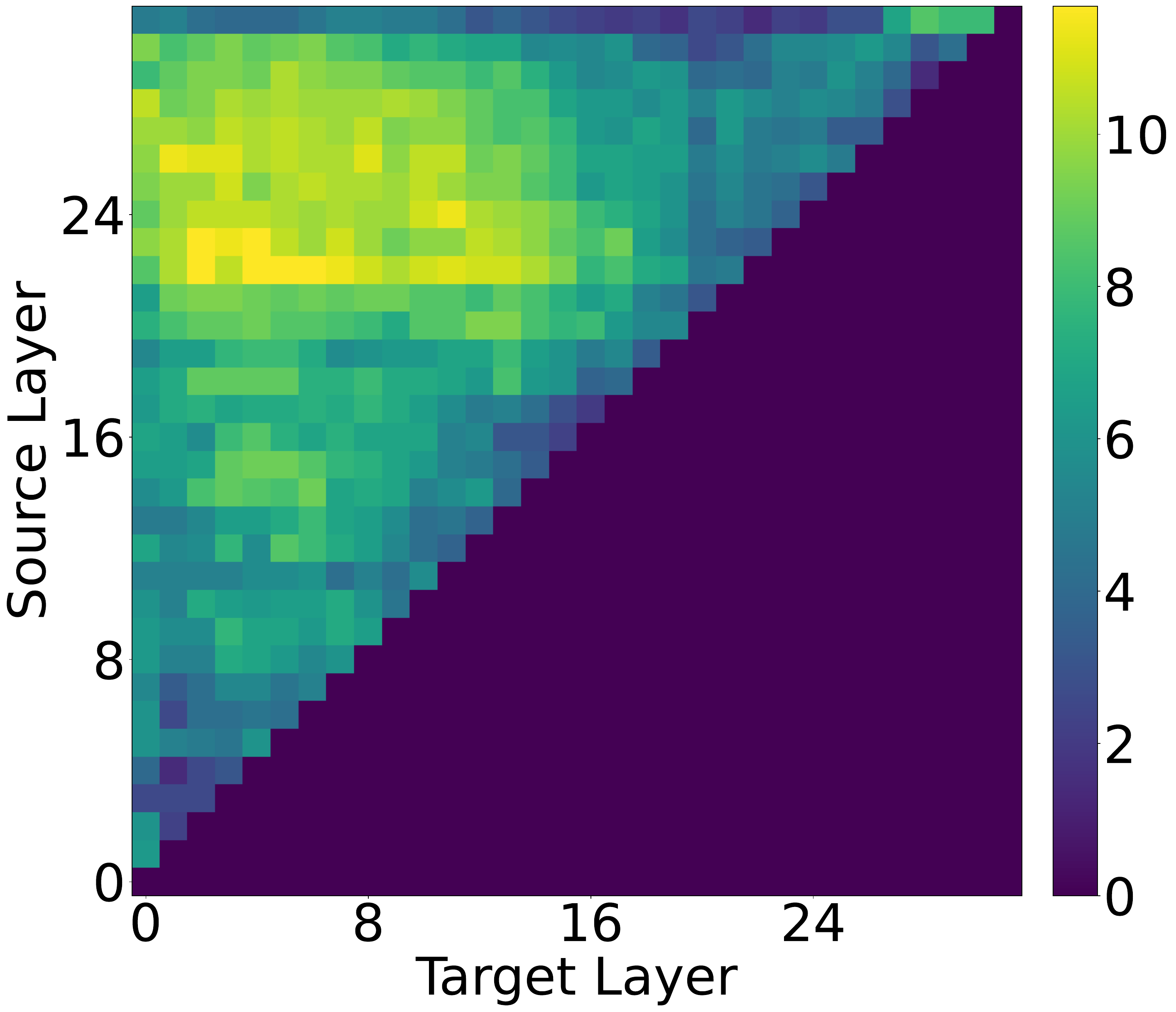}
    \caption{Source token is $t_2$, LLaMA 2 7B}    
    \label{fig:backpatching_last_llama2-7b}
\end{subfigure}
\par\bigskip
\begin{subfigure}{.5\textwidth}
    \centering
    \includegraphics[scale=0.12]{figures/backpatching_e1_llama2-13b.pdf}
    \caption{Source token is $t_1$, LLaMA 2 13B}    
    \label{fig:backpatching_e1_llama2-13b_app}
\end{subfigure}%
\hfill%
\begin{subfigure}{.5\textwidth}
    \centering
    \includegraphics[scale=0.12]{figures/backpatching_last_llama2-13b.pdf}
    \caption{Source token is $t_2$, LLaMA 2 13B}    
    \label{fig:backpatching_last_llama2-13b_app}
\end{subfigure}
\par\bigskip
\begin{subfigure}{.5\textwidth}
    \centering
    \includegraphics[scale=0.12]{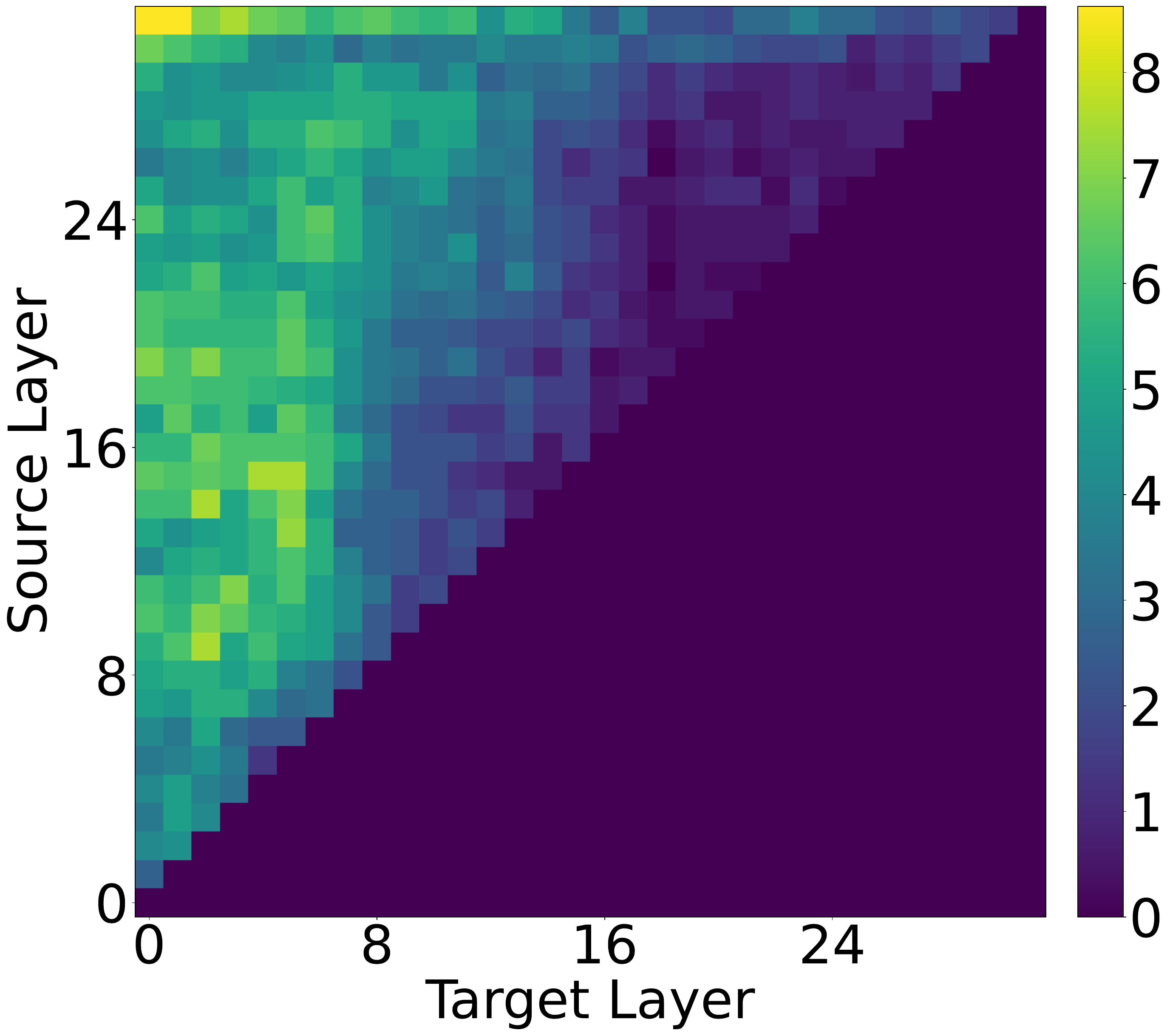}
    \caption{Source token is $t_1$, LLaMA 3 8B}    
    \label{fig:backpatching_e1_llama3-8b}
\end{subfigure}%
\hfill%
\begin{subfigure}{.5\textwidth}
    \centering
    \includegraphics[scale=0.12]{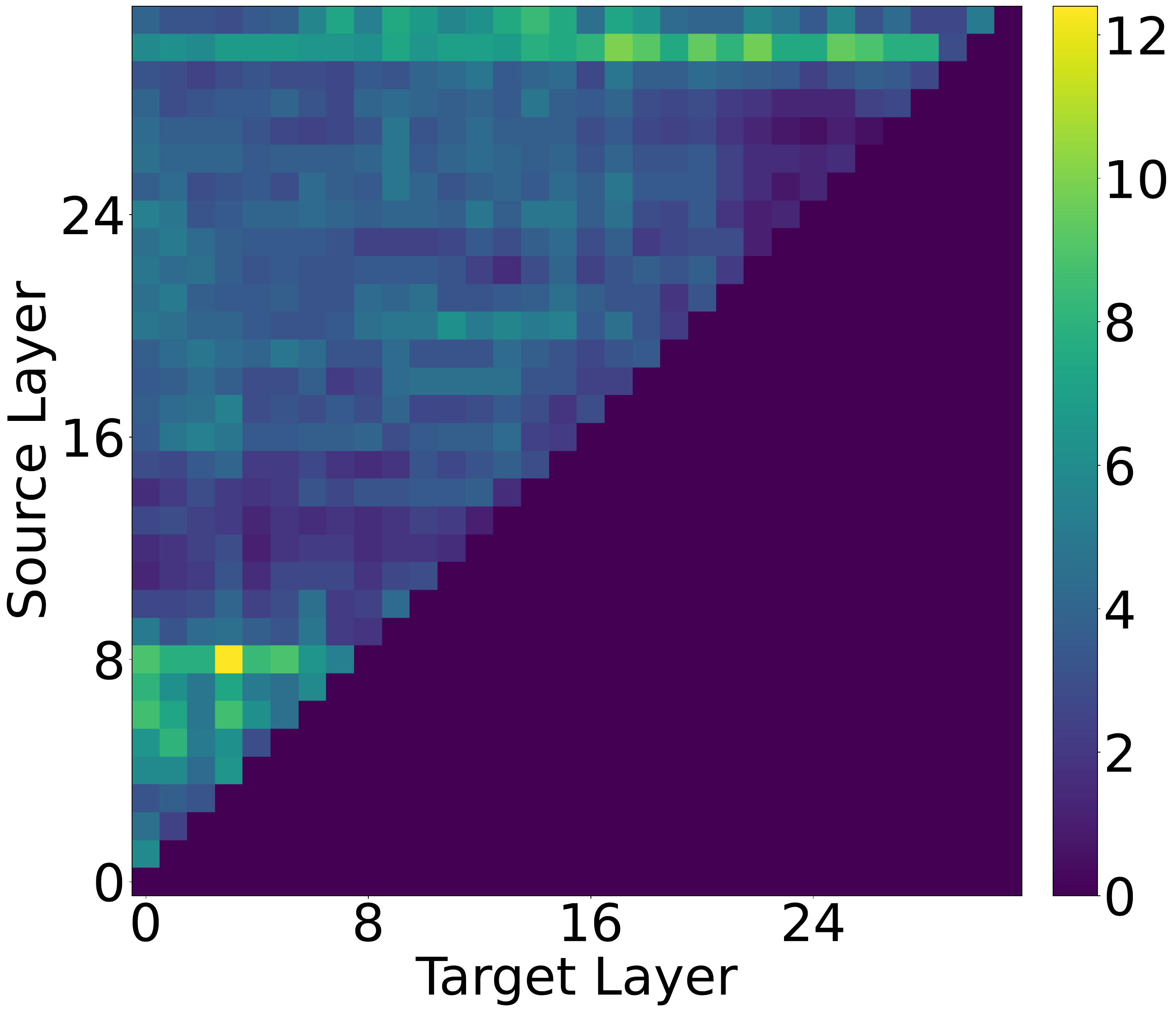}
    \caption{Source token is $t_2$, LLaMA 3 8B}    
    \label{fig:backpatching_last_llama3-8b}
\end{subfigure}
\par\bigskip
\begin{subfigure}{.5\textwidth}
    \centering
    \includegraphics[scale=0.12]{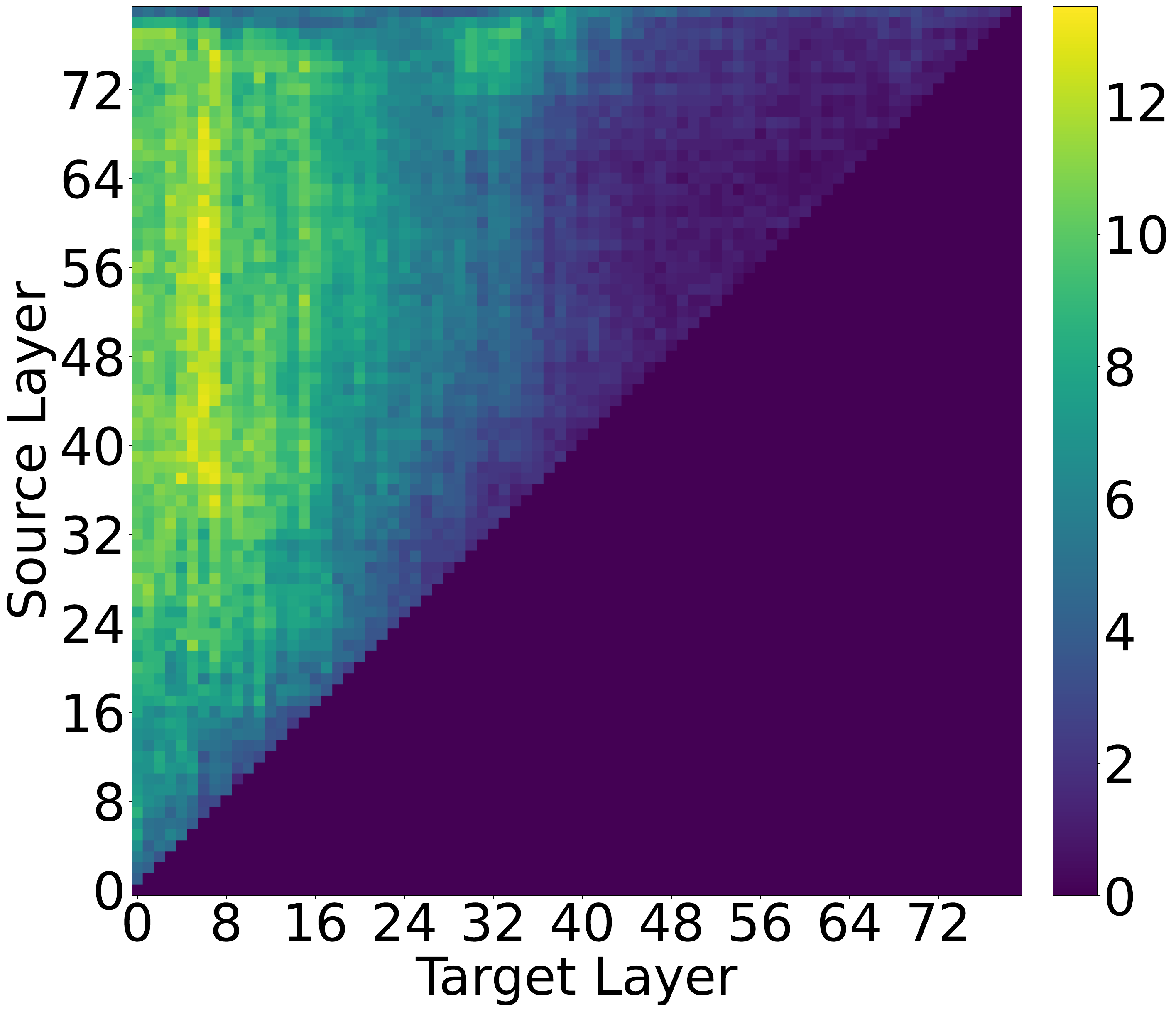}
    \caption{Source token is $t_1$, LLaMA 3 70B}    
    \label{fig:backpatching_e1_llama3-70b}
\end{subfigure}%
\hfill%
\begin{subfigure}{.5\textwidth}
    \centering
    \includegraphics[scale=0.12]{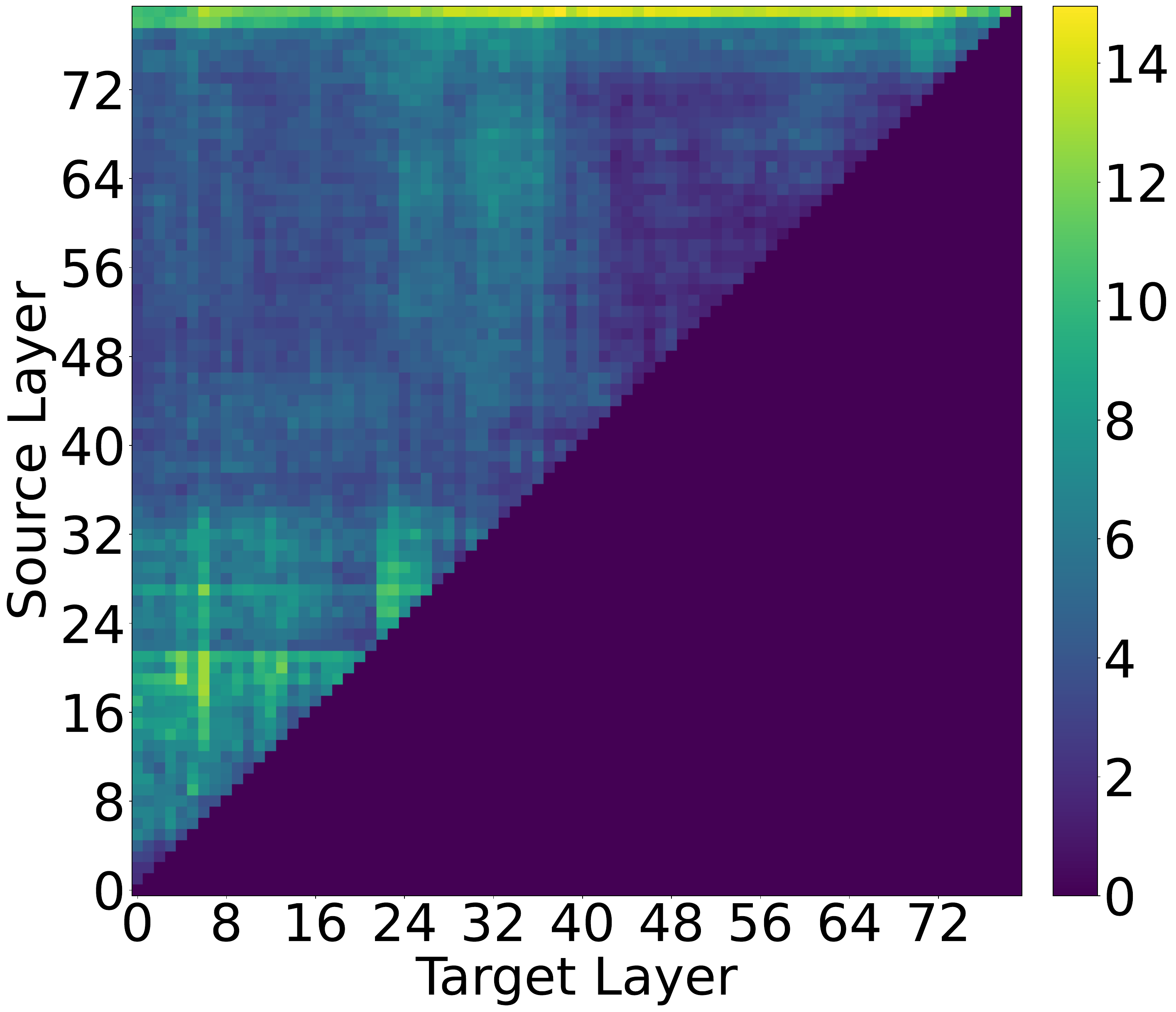}
    \caption{Source token is $t_2$, LLaMA 3 70B}    
    \label{fig:backpatching_last_llama3-70b}
\end{subfigure}
\caption{Back-patching heat-maps of successful source and target layers for LLaMA models.}
\label{fig:backpatching_llama_models}
\end{figure*}

\begin{figure*}[t]
\centering
\begin{subfigure}{.5\textwidth}
    \centering
    \includegraphics[scale=0.12]{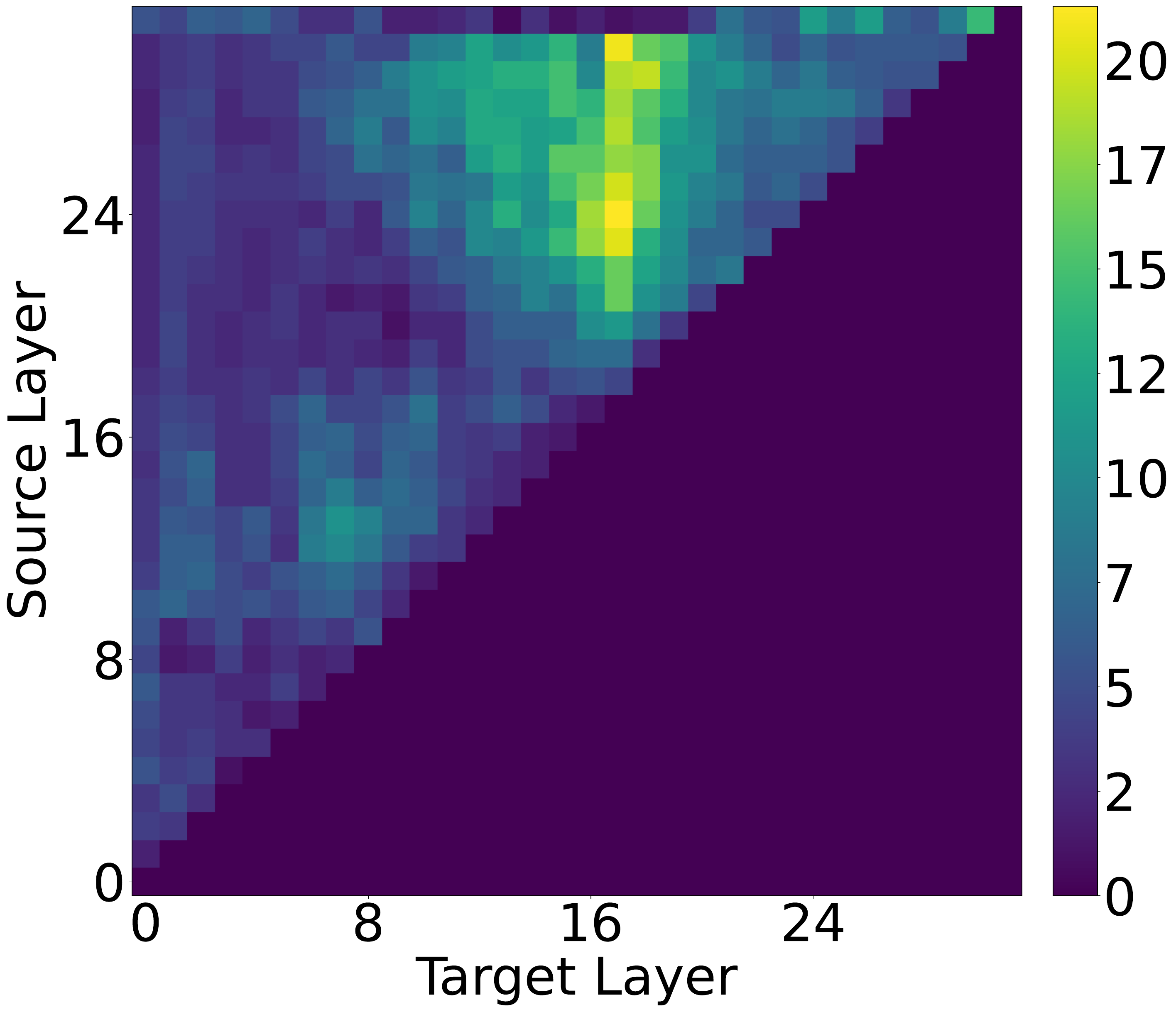}
    \caption{Source token is $t_1$, Pythia 6.9B}    
    \label{fig:backpatching_e1_pythia-6.9b}
\end{subfigure}%
\hfill%
\begin{subfigure}{.5\textwidth}
    \centering
    \includegraphics[scale=0.12]{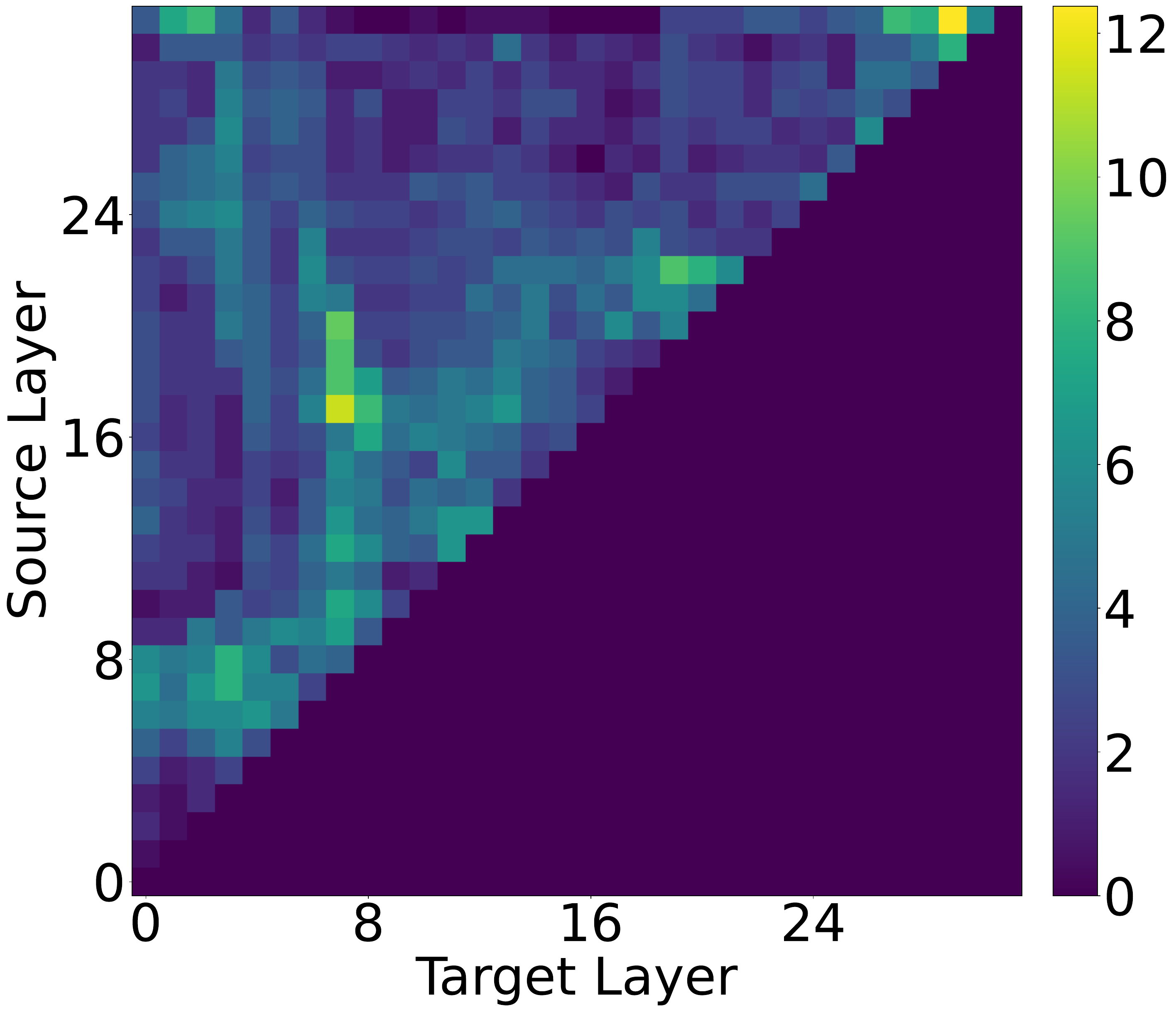}
    \caption{Source token is $t_2$, Pythia 6.9B}    
    \label{fig:backpatching_last_pythia-6.9b}
\end{subfigure}
\par\bigskip
\begin{subfigure}{.5\textwidth}
    \centering
    \includegraphics[scale=0.12]{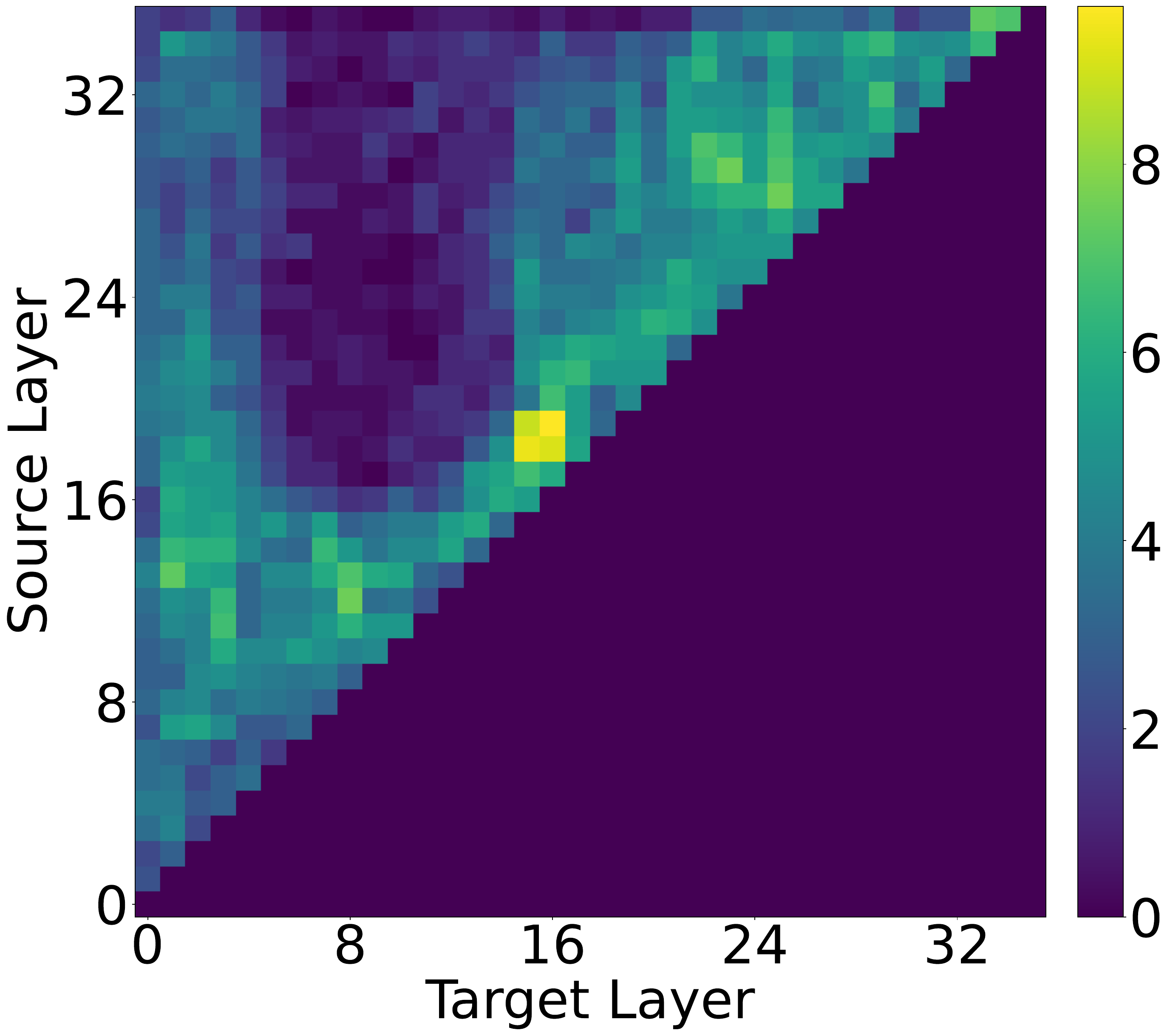}
    \caption{Source token is $t_1$, Pythia 12B}    
    \label{fig:backpatching_e1_pythia-12b}
\end{subfigure}%
\hfill%
\begin{subfigure}{.5\textwidth}
    \centering
    \includegraphics[scale=0.12]{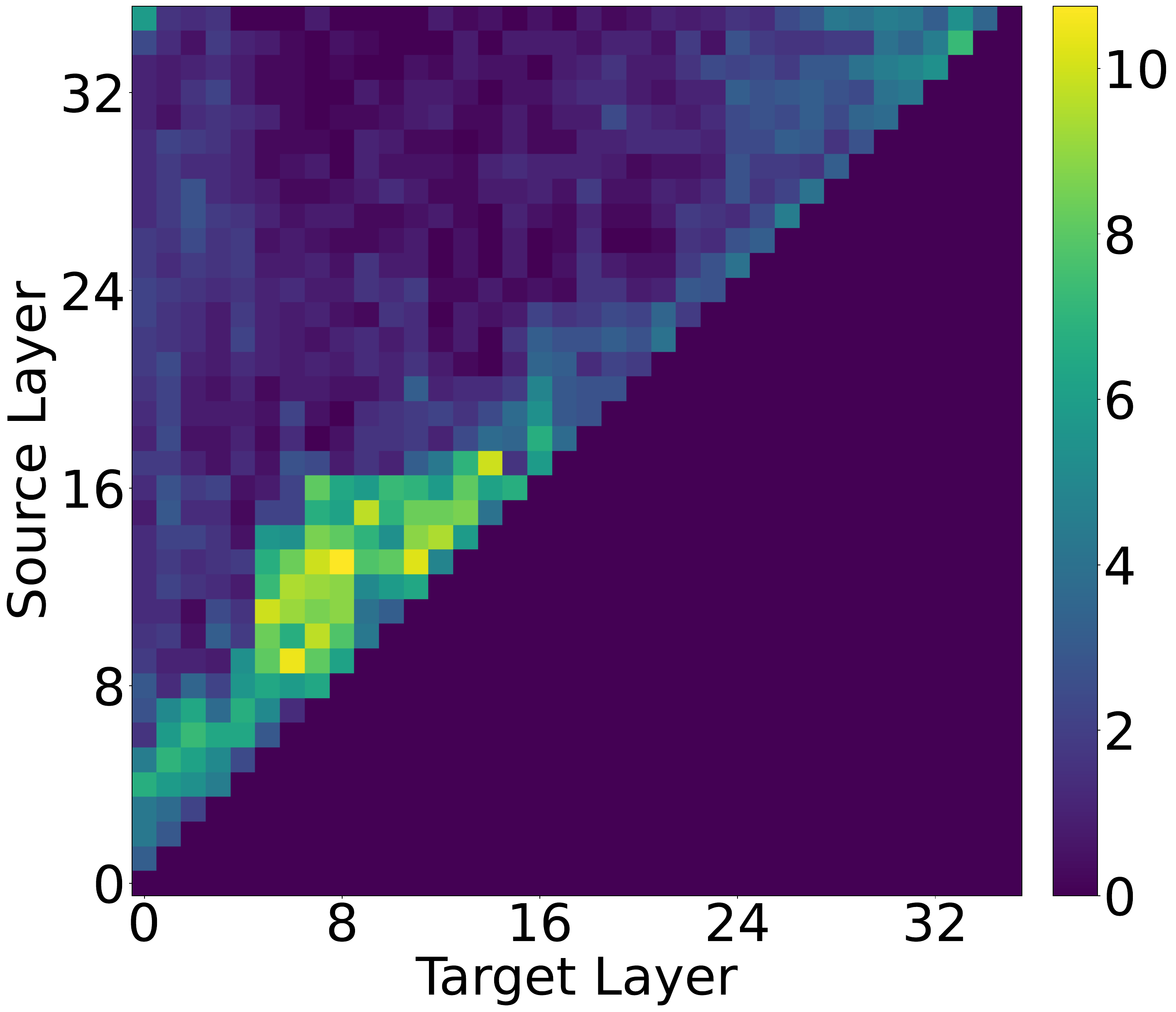}
    \caption{Source token is $t_2$, Pythia 12B}    
    \label{fig:backpatching_last_pythia-12b}
\end{subfigure}
\caption{Back-patching heat-maps of successful source and target layers for Pythia models.}
\label{fig:backpatching_pythia_models}
\end{figure*}

Figures \ref{fig:backpatching_llama_models} and \ref{fig:backpatching_pythia_models} depict the back-patching heap-maps of successful source and target layers (as shown in \S\ref{sec:backpatching}) for all models.

\end{document}